%% file: main.tex
\documentclass[runningheads]{llncs}

\usepackage[dvipsnames,table,xcdraw]{xcolor}
\usepackage[mobile]{eccv}

\usepackage{eccvabbrv}
\usepackage{graphicx}
\usepackage{booktabs}
\usepackage{amssymb}
\usepackage{multirow}
\usepackage{pgfplots}
\usepackage{pifont}
\usepackage{tikz}
\usepackage{xspace}
\usepackage{subcaption}
\usepackage{wrapfig}
\makeatletter
\renewcommand{\paragraph}{%
\@startsection{paragraph}{4}%
{\z@}{-0.3em}{-0.5em}%
{\normalfont\normalsize\bfseries}%
}
\makeatother
\usepackage[accsupp]{axessibility}  %
\newcommand{\method}{\emph{Lightplane}\xspace}
\newcommand{\renderer}{\emph{Renderer}\xspace}
\newcommand{\splatter}{\emph{Splatter}\xspace}
\newcommand{\PLH}{{\mkern-2mu\times\mkern-2mu}}

\usepackage[pagebackref,breaklinks,colorlinks,citecolor=eccvblue]{hyperref}

\begin{document}

\title{	
  Lightplane: Highly-Scalable Components for Neural 3D Fields}
\titlerunning{Lightplane: Highly-Scalable Components for Neural 3D Fields}

\author{Ang Cao\inst{1,2} \and
Justin Johnson\inst{2} \and
Andrea Vedaldi\inst{1} \and
David Novotny\inst{1}}

\authorrunning{A.Cao et al.}

\institute{Meta AI \and
University of Michigan}

\maketitle

\input{sec/0_abstract}

\input{sec/1_intro}
\input{sec/2_related}
\input{sec/3_method}
\input{sec/4_exp}

\input{sec/5_conclusion}

\clearpage 
\bibliographystyle{splncs04}
\bibliography{bib/main,bib/holo_diffusion, bib/holo_fusion, bib/vedaldi_general,bib/vedaldi_specific}

\appendix

\input{sec/supp_text}

\end{document}

%% file: sec/0_abstract.tex
\begin{abstract}
    Contemporary 3D research, particularly in reconstruction and generation, heavily relies on 2D images for inputs or supervision.
    However, current designs for these 2D-3D mapping are memory-intensive, posing a significant bottleneck for existing methods and hindering new applications.
    In response, we propose a pair of highly scalable components for 3D neural fields: \method  \renderer and \splatter, which significantly reduce memory usage in 2D-3D mapping.
    These innovations enable the processing of vastly more and higher resolution images with small memory and computational costs.
    We demonstrate their utility in various applications, from benefiting single-scene optimization with image-level losses to realizing a versatile pipeline for dramatically scaling 3D reconstruction and generation.
    Code: \url{https://github.com/facebookresearch/lightplane}.
\end{abstract}

%% file: sec/1_intro.tex
\input{figures/fig_teaser.tex}

\section{Introduction}%
\label{sec:intro}
Recent advancements in neural rendering and generative modeling have propelled significant strides in 3D reconstruction and generation.
However, in most cases, these methods are not exclusively 3D; instead, they heavily rely on 2D images as inputs or for supervision, which demands information mapping between 2D and 3D spaces.
For instance, Neural Radiance Fields~(NeRFs)~\cite{mildenhall2020nerf} use a photometric loss on 2D images rendered from 3D, bypassing direct 3D supervision. 
Similarly, various novel view synthesis and generation methods~\cite{chan2023genvs, yu2021pixelnerf, mvsnerf} employ 2D images as inputs and lift them into 3D space for further processing.
This mapping between 2D and 3D is critical for current 3D research, attributed to the scarcity of 3D training materials for developing versatile 3D models from scratch, and the relative ease of acquiring 2D images on a large scale.

Despite its crucial role and widespread use, the process of 2D-3D mapping incurs a high computational cost, especially in \emph{neural 3D fields} with volumetric rendering, which underpins many of the most powerful 3D representations.
These fields are defined by continuous functions that assign values, such as density or color, to \emph{any} point in 3D space, regardless of the presence of a physical surface. 
Therefore, they are powerful and flexible, preventing initialization in point rendering~\cite{kerbl3Dgaussians} or topology constraints for meshes~\cite{liu2019soft}.
The primary challenge lies in executing operations across numerous 3D points that span an entire volume.
While these operations can be relatively simple (\eg, evaluating a small multilayer perceptron (MLP) at each point, or extracting features from 2D input feature maps), performing them in a differentiable manner is \emph{extremely} memory intensive as all intermediate values must be kept in memory for backpropagation.

While the speed of NeRFs has been improved in~\cite{muller2022instant,Chen2022ECCV,fridovich2022plenoxels}, the issue of high memory consumption has seldom been studied.
This significant memory demand hampers scalability of 2D-3D communication, presenting a crucial bottleneck for many existing 3D models and a formidable barrier for potential new applications.
For example, the memory requirements to render even a single low-resolution image of a neural 3D field can be prohibitive enough to prevent the application of image-level losses such as LPIPS~\cite{zhang2018perceptual} or SDS~\cite{poole2022dreamfusion}.
Omitting such losses leads to a massive performance loss, as e.g. demonstrated by the state-of-the-art Large Reconstruction Model \cite{instant3d2023,hong2023lrm}.
Additionally, memory-inefficiencies limit the number of input images and the resolution of the 3D representation, preventing advancing from few-view novel view synthesis models~\cite{yu2021pixelnerf, Wang2021IBRNetLM} to large-scale amortized 3D reconstruction models with many conditioning images.

In this paper, we propose two highly scalable components for neural 3D fields: \method \renderer and \splatter.
These innovations enable 2D-3D mapping with four orders of magnitude less memory consumption while maintaining comparable speed.
\renderer renders 2D images of 3D models by means of the standard emission-absorption equations popularized by NeRF~\cite{mildenhall2020nerf}\@.
Conversely, \splatter lifts 2D information to 3D by splatting it onto the 3D representation,  allowing further processing with neural nets.
Both components are based on a hybrid 3D representation that combines `hashed' 3D representations such as voxel grids and triplanes with MLPs.
We use these representations as they are fast, relatively memory efficient, and familiar to practitioners, while components could be easily extended to other hashed representations as well.

As aforementioned, storing intermediate values at each 3D point for backpropagation causes tremendous memory usage.
We solve it by creatively reconfiguring inner computations and fusing operations over casted rays instead of 3D points.
Specifically, \method \renderer sequentially calculates features (\eg, colors) and densities of points along the ray, updating rendered pixels and transmittance on-the-fly without storing intermediate tensors.
This design significantly saves memory at the cost of a challenging backpropagation, which we solve by efficiently recomputing forward activations as needed.
Note that the latter is different from the standard ``checkpointing'' trick, whose adoption here would be of little help.
This is because checkpointing still entails caching many intermediate ray-point values as we march along each ray.

\method \splatter builds on similar ideas with an innovative design, 
where splatted features are stored directly into the hash structure underpinning the 3D model, without emitting one value per 3D point.
Besides voxel grids which are usually used for lifting, \splatter could be easily extended to other 3D hash structures. 
We implement these components in Triton~\cite{tillet2019triton}, a GPU programming language that is efficient, portable, and relatively easy to modify.
We will release the code as an open-source package upon publication.

Like convolution or attention, our components are designed as building blocks to boost a variety of 3D models and applications.
Empowered by the \method, we devise a pipeline taking up to input 100 images, significantly scaling the communication between 2D and 3D. 
We extensively evaluate on the CO3Dv2 dataset, reporting significant performance improvements in color and geometry accuracy for 3D reconstruction, and better 3D generation measured by FID/KID\@.
Finally, we boost performance of the state-of-the-art Large Reconstruction Model \cite{hong2023lrm}.

%% file: figures/fig_teaser.tex
\begin{figure}[!b]%
    \centering%
    \captionsetup{type=figure}%
   \includegraphics[width=1.0\linewidth,trim={0, 0, 0, 0},clip]{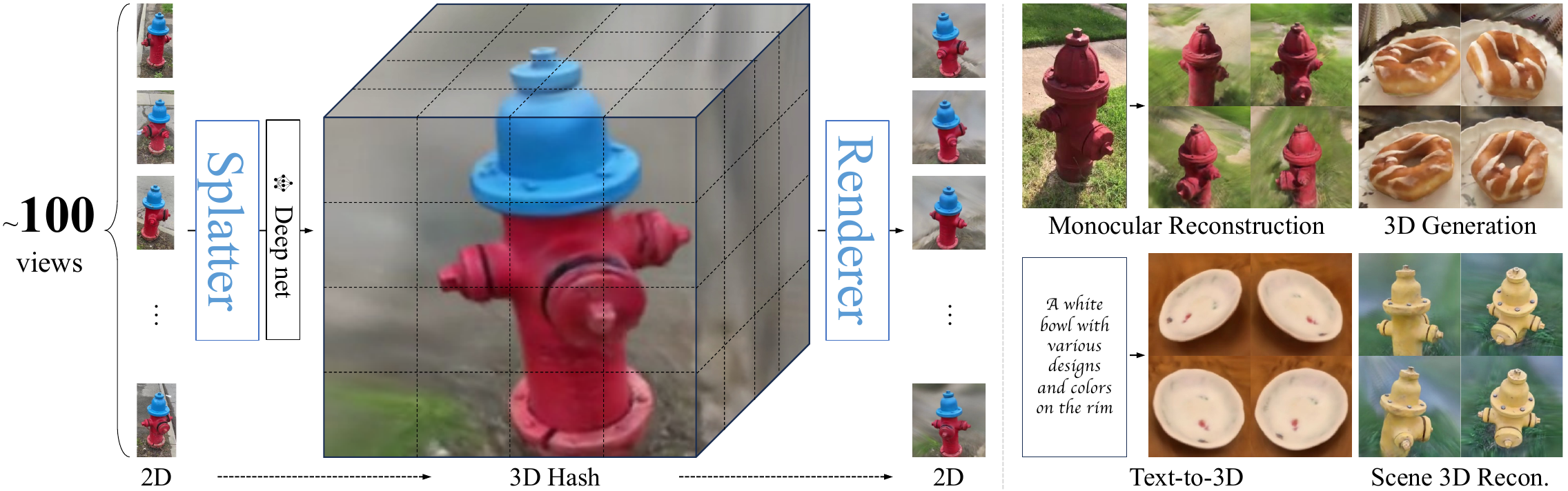}%
    \captionof{figure}{%
        We introduce the \textbf{\method \renderer} and \textbf{\splatter}, a pair of highly-scalable components for neural 3D fields (left).
        They address the key memory bottleneck of 2D-3D mapping (\ie rendering and lifting), and reduce memory usage by up to four orders of magnitude, 
        which dramatically increases the number of images that can be processed.
        We showcase how they can boost various 3D applications (right).
    }%
    \label{fig:teaser-fig}%
\end{figure}%

%% file: sec/2_related.tex
\section{Related Work}%
\label{sec:related}

\paragraph{3D reconstruction using neural 3D fields.}

Traditional 3D reconstruction models represented shapes as meshes~\cite{gkioxari19mesh,wang2018pixel2mesh},
point clouds~\cite{fan2017point,yang2019pointflow},
or voxel grids~\cite{choy20163d,girdhar2016learning}.
With the introduction of NeRF~\cite{mildenhall2020nerf}, however, the focus has shifted to \emph{implicit} 3D representations, often utilizing MLPs to represent occupancy and radiance functions defined on a 3D domain.
NeRF has been refined in many ways~\cite{barron2021mip,zhang2020nerf++,barron2022mip,verbin2022ref}, including replacing the opacity function with a signed-distance field to improve the reconstruction of surfaces~\cite{wang21neus:,wang2023neus2,li2023neuralangelo,rosu2023permutosdf,yariv2021volume,yu2022monosdf}.

Storing an entire scene in a single MLP, however, means evaluating a complex function anew at every 3D point, which is very expensive from both time and memory usage. 
Many authors have proposed to represent radiance fields with smaller, more local components to improve speed,
including using point clouds~\cite{xu22point-nerf:}, tetrahedral meshes~\cite{kulhanek2023tetra} or, more often, voxel grids~\cite{lombardi2019neural,yu2021plenoxels,sun2022direct,reiser21kilonerf:,karnewar2022relu}.
Voxel grids could be further replaced by more compact structures like low-rank tensor decompositions~\cite{chen22tensorf:}, triplanes~\cite{chan2022efficient}, hashing~\cite{muller22instant}, and their combination~\cite{reiser2023merf}.

Unlike the above methods focusing on speed, \method significantly reduces memory demands for neural 3D fields.
Note that our method targets neural 3D fields with volumetric rendering,  while point-based rendering like 3DGS~\cite{kerbl3Dgaussians} are not in this scope, since they don't model every 3D point in the space and rely on rasterization instead of volumetric rendering.
While 3DGS exhibits fast convergence speed, importantly, it has been shown to give lower accuracy (measured in PSNR) in both the single-scene overfitting case \cite{barron2023zip}, and the few-view reconstruction case \cite{tochilkin24triposr}.
For optimal performance in single-scene, they require careful surface initialization whereas NeRFs converge from a random initialization.

\paragraph{Amortized 3D reconstruction.}
Amortized~(Generalizable) 3D reconstruction utilizing implicit shape representations was initially approached in~\cite{yu2021pixelnerf,henzler21unsupervised,grf2020,reizenstein21common,wang21ibrnet:,nguyen2021rgbd} by warping/pooling features from source views to a target to estimate the color of the underlying scene surfaces.
\cite{wu2023multiview,sajjadi21scene} introduces latent transformer tokens to support the reconstruction.
Generalizable triplanes~\cite{irshad2023neo, hong2023lrm, li2023instant3d}, ground-planes~\cite{sharma22seeing}, and voxel grids~\cite{kar2017learning} were also explored.

A common downside of these methods is their memory consumption which limits them all to a \emph{few-view} setting with up to 10 source views.
They either are trained on a category-specific dataset or learn to interpolate between input views with unsatisfactory geometry and 3D consistency.
Owing to its memory efficiency, \method allows more than 100 input source views.
We leverage the latter to train a large-scale 3D model yielding more accurate reconstructions.

\paragraph{Image-supervised 3D generators.}

With the advent of Generative Adversarial Networks~\cite{goodfellow14generative} (GAN), many methods attempted to learn generative models of 3D shapes given large uncurated image datasets.
PlatonicGAN~\cite{henzler19escaping}, HoloGAN~\cite{nguyen-phuoc19hologan:} and PrGAN~\cite{gadelha163d-shape} learned to generate voxel grids whose renders were indistinguishable from real object views according to an image-based deep discriminator.
The same task was later tackled with Neural Radiance Fields~\cite{schwarz2020graf,chan2022efficient,gu2021stylenerf,skorokhodov2022epigraf,Niemeyer2020GIRAFFE},
and with meshes~\cite{gao22get3d:,wu2020unsupervised}.
The success of 2D generative diffusion models~\cite{dhariwal2021diffusion} led to image-supervised models such as HoloDiffusion~\cite{karnewar2023holodiffusion}, Forward Diffusion~\cite{tewari2023forwarddiffusion}, and PC$^2$~\cite{melaskyriazi2023projection}, which directly model the distribution of 3D voxel grids, implicit fields and point clouds respectively.
Similarly, RenderDiffusion~\cite{anciukevicius22renderdiffusion:} and ViewsetDiffusion~\cite{szymanowicz2023viewset} learn a 2D image denoiser by means of a 3D deep reconstructor.
GeNVS~\cite{chan2023genvs} and HoloFusion~\cite{karnewar2023holofusion} proposed 3D generators with 2D diffusion rendering post-processors.
We demonstrate that \method brings a strong performance boost to ViewsetDiffusion and generates realistic 3D scenes.  

%% file: sec/3_method.tex
\section{Method}%
\label{sec:method}

\newcommand{\ba}{\boldsymbol{a}}
\newcommand{\bbb}{\boldsymbol{b}}
\newcommand{\x}{\boldsymbol{x}}
\newcommand{\br}{\boldsymbol{r}}
\newcommand{\bv}{\boldsymbol{v}}
\newcommand{\bp}{\boldsymbol{p}}

We introduce the \method \renderer and  \splatter, which facilitate the mapping of information between 2D and 3D spaces in a differentiable manner, significantly reducing memory usage in the process. 
We first discuss the memory bottlenecks of existing methods that are used for rendering and lifting images into 3D structures (\cref{sec:method_preliminary}). 
Then we define the hashed 3D representations~(\cref{sec:method_3DRep}) used in our framework and functionality~(\cref{sec:method_function}) of the proposed components. 
Lastly, we discuss their implementations~(\cref{sec:memory_efficient_implementation}).

 \subsection{Preliminary}
 \label{sec:method_preliminary}

\paragraph{2D-3D Mapping.}
Mapping between 2D images and 3D models is a major practical bottleneck of many algorithms (\cref{sec:intro}), particularly when using powerful implicit 3D representations such as neural 3D fields.
The memory bottleneck comprises a large number of 3D points from rendering rays and their intermediate features, which are cached in GPU memory for the ensuing backpropagation.

\input{figures/fig_components.tex}

More specifically, for rendering~(\emph{3D to 2D mapping}), an \emph{entire ray} of 3D points contributes to the color of a single pixel in the rendered image.
With $M$ pixels and $R$ points per ray, $M{\PLH}R$ implicit representation evaluations are required to get 3D points' colors and opacities.
All these intermediate results, including outputs of all MLP layers for every 3D point, are stored in memory for backpropagation, leading to huge memory usage. 
Using a tiny MLP with $L{=}6$ layers and $K{=}64$ hidden units, $M\PLH R \PLH L \PLH K$ memory is required to \emph{just} store the MLP outputs, which totals 12 GB for a $256^2$ image with $R{=}128$ points per ray.

Similarly, to lift $N$ input features to 3D ~(\emph{2D to 3D mapping}), popular models like PixelNeRF~\cite{yu2021pixelnerf} and GeNVS~\cite{chan2023genvs} project each 3D point to $N$ input views individually, and average $N$ sampled feature vectors as the point feature.
Even without considering any MLPs, $N \PLH |\mathcal{M}|$ memory is used, where $|\mathcal{M}|$ is the size of 3D structure $\mathcal{M}$.
When $\mathcal{M}$ is a $128^3$ voxel grid with 64-dimensional features, $|\mathcal{M}|$ takes 512 MB in FP32, leading to 5 GB of memory with just 10 input views.

Moreover, the aforementioned lifting requires 3D positions for projection and cannot be easily generalized to other compact representations like triplanes, since cells in such ``hashed'' feature maps (\eg 2D position on feature planes for triplanes) don't have clearly-defined 3D positions. 
Hence, directly lifting multi-view features to triplanes for further processing is still an open problem.

The memory bottleneck impacts several aspects.
For mapping from 3D to 2D (\ie rendering), methods like NeRF~\cite{mildenhall2020nerf} and  PixelNeRF~\cite{yu2021pixelnerf} are limited to a few low-resolution images per training iteration (even using 40GB GPUs) or to subsample rendered pixels, which prohibits image-level losses such as LPIPS~\cite{zhang2018perceptual} and SDS~\cite{poole2022dreamfusion}.
For mapping from 2D to 3D, memory demands limit input view numbers and 3D representation sizes.
The huge memory usage not only occupies resources that could otherwise enhance model sizes and capacities but also restricts model training and inference on devices with limited memory availability.

\paragraph{Neural 3D fields.}
Let $\x \in \mathbb{R}^3$ denote a 3D point, a \emph{neural 3D field} is a volumetric function $f$ that maps each point $\x$ to a vector $f(\x) \in \mathbb{R}^C$.
NeRF~\cite{mildenhall2020nerf} represented such functions using a single MLP.
While this is simple, the MLP must represent \emph{whole} 3D objects and hence must be large and costly to evaluate.

Several approaches are proposed to solve this problem, by decomposing information into local buckets, accessing which is more efficient than evaluating the global MLP\@.
Most famously, \cite{muller2022instant} utilizes hash tables, but other representations such as voxel grids~\cite{SunSC22} and various low-rank decompositions such as triplanes~\cite{chan2022efficient}, TensoRF~\cite{chen2022tensorf} and HexPlane~\cite{Chan2021, cao2023hexplane} also follow this pattern.

\subsection{Hybrid representation with 3D hash structure}
\label{sec:method_3DRep}

Following the idea in \cref{sec:method_preliminary}, we use a hybrid representation for neural 3D fields $f$, and decompose $f = g \circ h$, where $h : \mathbb{R}^3 \rightarrow \mathbb{R}^{K}$ is a hashing scheme~(sampling operation) for 3D hash structure $\theta$, and $g : \mathbb{R}^{K} \rightarrow \mathbb{R}^C$ is a tiny MLP, which takes features from hashing as inputs and outputs the final values.
In this paper, we generalize the concept of 3D hash structures to structures like voxel grids~\cite{SunSC22}, triplanes~\cite{chan2022efficient}, HexPlane~\cite{Chan2021, cao2023hexplane} and actual hash table~\cite{muller2022instant}, as obtaining information from these structures only requires accessing and processing the small amount of information stored in a particular bucket.
The associated hashing scheme $h$ typically samples 3D point features from hash structure $\theta$ via interpolation, which is highly efficient. 
In practice, we operationalize $\theta$ with voxel grids and triplanes as they are easy to process by neural networks, although other structures with a differentiable hashing scheme could be easily supported. 

In more detail, in the voxel-based representation, $\theta$ is a $H\PLH W\PLH D\PLH K$ tensor and $h$ is the tri-linear interpolation on $\theta$ given position $\x$.
In the triplane representation, $\theta$ is a list of three tensors of dimensions
$H\PLH W\PLH K$,
$W\PLH D\PLH K$, and
$D\PLH H\PLH K$.
Then, $h(x,y,z)$ is obtained by bilinear interpolation of each plane at
$(x,y)$,
$(y,z)$,
$(z,x)$, respectively, followed by summing the resulting three feature vectors.
Again, this design could be easily generalized to other hashed 3D structures $\theta$ and their corresponding hashing scheme~(sampling operation) $h$. 

\subsection{Rendering and splatting}
\label{sec:method_function}
We now detail \method \renderer and \splatter, two components using hybrid 3D representations with 3D hash structures. 
They are mutually dual %
as one maps 3D information to 2D via rendering, and the other maps 2D images to 3D.

\input{figures/figure_splatter}

\paragraph{\renderer.}
\renderer outputs pixel features $\bv$~(\eg colors, depths) in a differentiable way from a hybrid representation $f{=}g\circ h$, given $M$ rays $\{\br_i\}_{i=1}^{M}$ and $R{+}1$ points per ray.
We make its high-level design consistent with existing hybrid representations
\cite{SunSC22,chan2022efficient,cao2023hexplane,muller2022instant}
as they have proven to be powerful, while re-designing the implementation in \cref{sec:memory_efficient_implementation} to achieve significant memory savings.

Following volumetric rendering of NeRF~\cite{mildenhall2020nerf}, \renderer uses a generalized Emission-Absorption~(EA) model and calculates transmittance $T_{ij}$, which is the probability that a photon emitting at $\x_{ij}$ ($j$-th sampling points on the $i$-th ray) reaches the sensor. 
Accordingly, the rendered feature $\bv_i$ of ray $\br_{i}$ is:
\begin{equation}\label{e:render}
    \bv_i
    =
    \sum_{j=1}^{R} (T_{i,j-1} - T_{ij}) f_v(\x_{ij}).
\end{equation}
where $f_v(\x_{ij})$ is the feature~(\eg color) of the 3D point $\x_{ij}$, obtained from the hybrid representation $f_v$; $T_{ij} = \exp(-\sum_{n=0}^{j}\Delta \cdot \sigma(\x_{in}))$,
$\Delta$ is the distance between two sampled points, and $\sigma(\x_{in})$ is the opacity of the $n$-th sampled point;
$(T_{i, j-1} - T_{ij}) \in [0, 1]$ is the visibility of the point $\x_{ij}$.
Given a 3D point, \renderer samples its feature from the 3D representation and feeds the feature to an MLP $g_{\sigma}$ to calculate the opacity. 
$f_v(\x_{ij})$ is calculated by another MLP $g_v$ taking the sampled feature and view directions as inputs.

\paragraph{\splatter.}
Opposite to \renderer, \splatter maps input view features to 3D hash structures.
Existing works like \cite{chan2023genvs, szymanowicz2023viewset,karnewar2023holodiffusion,karnewar2023holofusion} achieve this by looping over all points \emph{inside voxel grids} and \emph{pulling} information from input features.
They project 3D points to input views, interpolate fields of 2D image features, compute and store a feature vector for each 3D sample.
Such operations are inherently memory-intensive and cannot be easily generalized to other 3D hash structures~\cref{sec:method_preliminary}.

Instead of looping over 3D points and pulling information from inputs, we make \splatter loop over \emph{input pixels/rays} and directly \emph{push} information to 3D structures. 
This makes \splatter a reversion of \renderer, being able to easily extend to other 3D structures and enjoy similar memory optimization designs.

Given $M$ input pixels, \splatter expands each pixel into a ray $\br_i$ with $R+1$ equispaced 3D points $\x_{ij}$, with points along the ray inheriting the pixel's features $\bv_i$.
3D points' features $\bv_{ij}$ are splatted back to zero-initialized 3D structures $\theta$, which operation is inverse to the sampling operation $h(\x)$ used in rendering.
This is done by accumulating $\bv_{ij}$ to hash cells that contain $\x_{ij}$, which accumulation is weighted by splatting weights.
After accumulating over all $M$ rays, each hash cell is normalized by the sum of all splatting weights landing in the cell.
The splatting weights are the same as the sampling weights used in rendering.
For voxel grids, a hash cell is a voxel, and splatting weights are the normalized inverse distance between the 3D point and eight voxel vertices. 
It can be easily extended to other hash structures.
We illustrate this splatting operation in Figure~\ref{fig:splat_figure}(b).

This na\"ive version of \splatter works well for voxel grids, but fails to work on triplanes and potentially other hashed representations.
We hypothesize it is due to 3D position information being destroyed when reducing from 3D space to 2D planes, and accumulated features are unaware of the spatial structure of the 3D points.
To address this, we propose to use an MLP $g_s$ to predict a modified feature vector $\bv_{ij}$ from the input vectors $\bv_i$, interpolated prior shape encoding $h_{\hat{\theta}}(\x_{ij})$, and the positional encoding $\operatorname{direnc}(\br_{i})$ of ray direction $\br_{ij}$.
For each sample $\x_{ij}$, the splatted feature $\tilde{\bv}_{ij}$ is
\begin{equation}
    \tilde{\bv}_{ij} = g_s(\bv_{i},h_{\hat{\theta}}(\x_{ij}),\operatorname{direnc}(\br_{ij}))
\end{equation}
$\hat{\theta}$ is a another hashed 3D representation, where prior shape encoding of 3D point $\x_{ij}$ could be obtained by hashing operation $h_{\hat{\theta}}(\cdot)$.
This MLP allows points along the same ray to have different spatial-aware features and thus preserves the spatial structure of the 3D points.
This design also allows us to iteratively refine 3D representations $\theta$ based on previous representations $\hat{\theta}$ and input features.

\subsection{Memory-efficient Implementation}%
\label{sec:memory_efficient_implementation}
We discuss the practical implementations of \method \renderer and \splatter, which are designed to be memory-efficient and scalable.

\paragraph{Fusing operations along the ray.}
As analyzed, current rendering and lifting operations for neural 3D fields are memory intensive, as they treat 3D points as basic entities and store intermediate results for each point.
Alternatively, we treat rays as basic entities and fuse operations in a single GPU kernel, where each kernel instance is responsible for a single ray.
This allows us to only store the rendered features and accumulated transmittance of the ray.

As Eq.~\ref{e:render}, a \renderer kernel sequentially samples 3D points' features, calculates features and opacities via MLPs and updates the rendered results and accumulated transmittance of the ray. 
These processes are integrated into a single kernel, obviating the need for storing any other intermediate results.
For the example in \cref{sec:method_preliminary}, memory usage is significantly reduced from $O(MKRL)$ to $O(MK)$, decreasing from 12 GB to 2 KB for an image of size $256^2$ with $R=128$ samples per ray in FP32.
This is less than 0.02\% of the memory required by the na{\"\i}ve implementation.
Since \splatter is designed to process rays emanating from input pixels as well~(\cref{sec:method_function}), it benefits from the same optimization practice.

\paragraph{Recalculation for backpropagation.}
Saving \emph{no} intermediate results during forward propagation significantly decreases memory usage, while these tensors are essential for backpropagation.
To solve it, we recompute the intermediate results during backpropagation for gradient calculation.
Speed-wise, recalculating the MLP in the forward direction increases the total number of floating-point operations by less than 50\% compared to the na{\"\i}ve implementation.
But this cost only occurs during backpropagation, and leads to massive memory savings.

\paragraph{Leveraging GPU memory hierarchy for speed.}
The speed could be further optimized by exploiting the hierarchical architecture of GPU memory.
By fusing operations in a single GPU kernel, we enhance the utilization of GPU's on-chip SRAM, and prevent massive access to GPU's high bandwidth memory (HBM).
Given that HBM access speeds are substantially slower compared to on-chip SRAM, and performance bottlenecks often stem from HBM access during tensor read/write operations, our kernel maintains a competitive speed even while recalculating intermediate results for backpropagation.
We encourage readers to refer to flash-attention~\cite{dao2022flashattention} for details of GPU memory hierarchy.

\paragraph{Emission-absorption backpropagation.}
\renderer and \splatter are dual to each other not only in functionality but also in their high-level implementation.
The backpropagation process of \splatter mirrors the forward pass of \renderer, as it samples 3D point gradients from the representation's gradient field and aggregates them along the ray to form the input pixel's gradients.
Conversely, \renderer's backward process is also similar to \splatter's forward pass.

\input{figures/fig_kernel_benchmark.tex}

Notably, the backpropagation of \renderer is more complicated as the visibility of 3D points is affected by the transmittance of previous points in the emission-absorption model. 
During forward pass, we sequentially calculate 3D points' visibility and implement the rendering equation \cref{e:render} by summing in order $j=1,2,\dots$, as it is easy to obtain visibility $T_j$ from $T_{j-1}$~(we omit ray index for simplicity).
For backward pass, on a ray $\br$, we derive the vector-Jacobian product (\ie, the quantity computed during backpropagation) of \renderer:
\begin{align}
\bp^\top \frac{d\bv}{df_\sigma(\x_{q})}
&=
- \Delta \frac{d \sigma(\x_{q})}{df_\sigma(\x_{q})}
\left(
\sum_{j=q+1}^R (T_{j-1} - T_j)\ba_j - T_q \ba_q
\right),\label{e:back}
\end{align}
where $\ba_j = \bp^\top f_v(\x_{j})$ and  $\bp$ is the gradient vector that needs backpropagating.

To backpropagate through  \renderer efficiently, we compute \cref{e:back} by marching along each rendering ray in the reverse order $q=R,R-1,\dots$, since the vectors $\ba_j$ are accumulated from sample $q$ onwards, and the opacity $f_\sigma(\x_q)$ affects only the visibility of successive samples $\x_q,\x_{q+1},\ldots$.
To make this possible, we cache the final transmittance $T_{R}$, which is computed in the forward pass(this amounts to one scalar per ray).
In backpropagation, we sequentially compute $\sigma(x_j)$ for every 3D point along the ray, and calculate $T_{j-1} = T_j \cdot \exp(\Delta \sigma(x_j))$ from $T_j$.
This way, similar to the forward pass, the kernel only stores the accumulation of per-point features instead of keeping them all in memory.

\paragraph{Difference from checkpointing.} Note that the latter is very different from ``checkpointing'' which can be trivially enabled for the naive renderer implementation in autograd frameworks such as PyTorch.
This is because, unlike our memory-efficient backward pass from \cref{e:back}, a checkpointed backward pass still entails storing all intermediate features along rendering rays in memory.

\section{Example applications}%
\label{sec:rec}
We show various 3D applications that could be boosted by the proposed components, from single-scene optimization with image-level losses to a versatile framework for large-scale 3D reconstruction and generation.
Results are in \cref{sec:exp}.

\paragraph{Single-scene optimization with image-level losses.}
Constrained by intensive memory usage during rendering, existing volumetric methods are limited to optimizing pixel-level losses on a subset of rays, such as MSE, or using image-level losses on low-resolution images ($64\times64$).
In contrast, we show how \renderer allows seamless usage of image-level losses on high-resolution renders.

\paragraph{Multi-view reconstruction.}
Combining \renderer and \splatter, we introduce a versatile pipeline for 3D reconstruction and generation. 
Given a set of views (viewset) $\mathcal{V}{=}\{I_i\}_{i=1}^{N}$ and corresponding cameras $\{\pi_i\}_{i=1}^N$, we train a large-scale model  $\Phi$,
which directly outputs the 3D representations $\theta{=}\Phi(\mathcal{V},\pi)$ of the corresponding scene by learning 3D priors from large-scale data.
Reconstruction starts by extracting a pixel-wise feature map $\bv{=}\psi(I_i)$ from each image $I_i$ and lifting them into the 3D representation $\tilde \theta$ with \splatter. 
Model $\Phi$ takes $\tilde \theta$ as input and outputs the final 3D representations $\theta {=}\Phi_\theta(\tilde \theta)$. 
Finally, \renderer outputs novel view images $\hat I = \mathcal{R}(\theta,\pi)$ from $\theta$, and the model is trained by minimizing the loss $\mathcal{L}$ between the novel rendered image and the corresponding ground truth $I$.

\paragraph{3D generation using viewset diffusion.}
Following recent works~\cite{szymanowicz2023viewset,anciukevivcius2023renderdiffusion}, this 3D reconstruction pipeline could be extended into a diffusion-based 3D generator with very few changes.
This is achieved by considering a noised viewset as input to the network, and training the model to denoise the viewset,  where each image $I_i$ is replaced with $I_{it} = \alpha_t I_i + \sigma_t \epsilon_i$ where $t$ is the noising schedule, $\alpha_t$ and $\sigma_t = \sqrt{1-\alpha_t^2}$ are the noise level, and $\epsilon_i$ is a random Normal noise vector.
During inference, the model initializes the viewset with Gaussian noise and iteratively denoises by applying the reconstruction model.
This process simultaneously generates multiple views of the object as well as its 3D model.

%% file: figures/fig_components.tex
\begin{wrapfigure}{r}{0.6\textwidth}%
\vspace{-0.4cm}%
\centering%
  \includegraphics[width=1.0\linewidth]{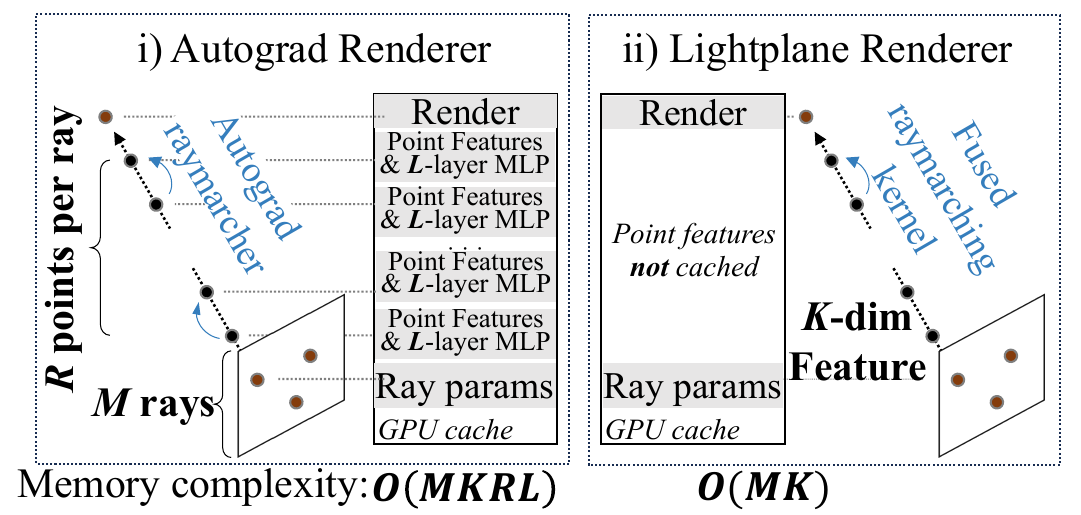}%
    \vspace{-0.1cm}%
    \caption{%
        \textbf{Memory usage} of our \method \renderer vs. a standard autograd NeRF renderer.%
    }\label{fig:components}%
\vspace{-0.4cm}%
\end{wrapfigure}

%% file: figures/figure_splatter.tex
\begin{figure}[!th]
\centering
\includegraphics[width=0.95\linewidth]{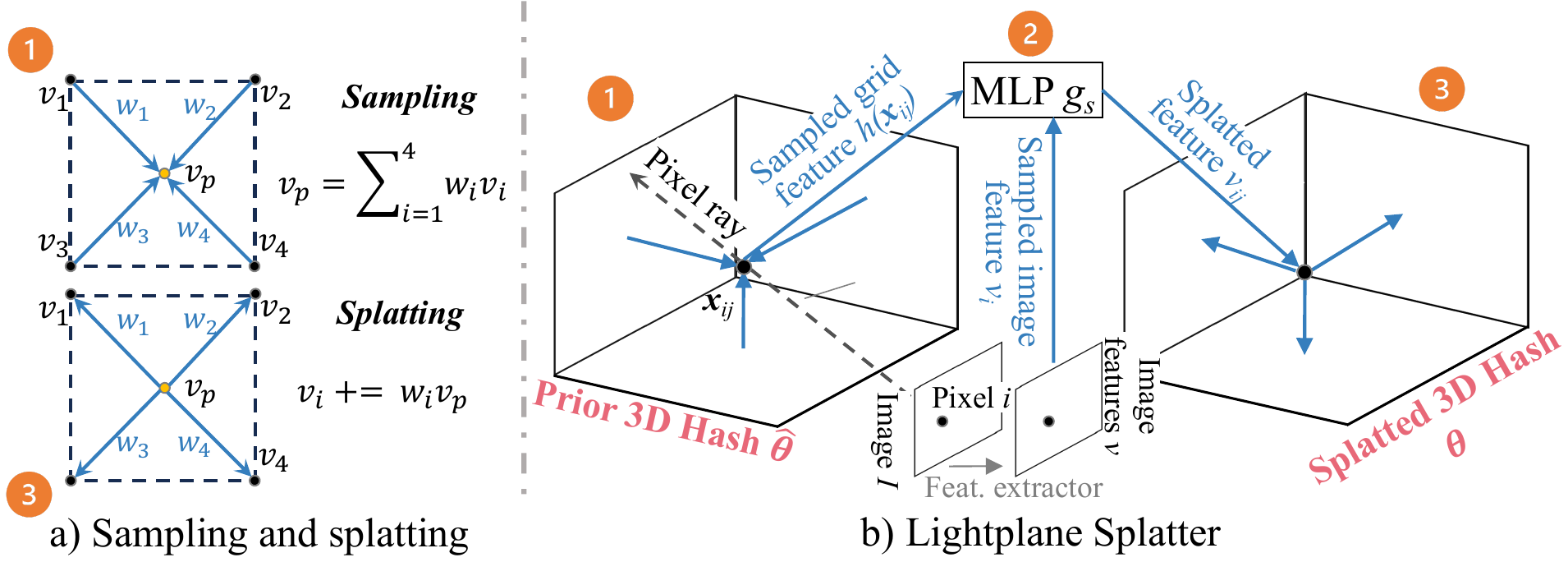}
\vspace{-3mm}
\caption{\textbf{\method \splatter.}
(a) On a hash grid with vertex features ${\bv_i}$: \emph{sampling} obtains point features $\bv_{p}$ by interpolating vertex features weighted by inverse distance; \emph{splatting} updates vertex features by accumulating point feature to vertex using the same weights.
(b) \splatter involves three steps. 
For each 3D point along the ray, \splatter samples its features from prior 3D hash $\hat{\theta}$ (1), calculates features to be splatted using  MLP (2), and splats them to zero-initialized $\theta$ (3). 
}
\label{fig:splat_figure}
\vspace{-4mm}  %
\end{figure}

%% file: figures/fig_kernel_benchmark.tex
\begin{figure}[!t]
    \centering
    \definecolor{tritonfw}{RGB}{137, 207, 240}
    \definecolor{tritonbw}{RGB}{137, 207, 240}
    \definecolor{pytorchfw}{RGB}{96, 130, 182}
    \definecolor{pytorchbw}{RGB}{96, 130, 182}
    \includegraphics[width=1.0\textwidth]{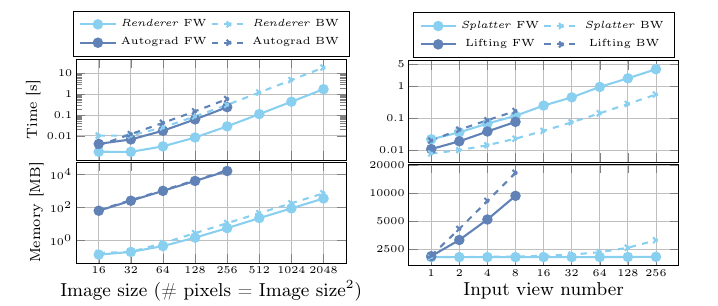}
    \caption{\textbf{\method memory \& speed benchmark} showing the forward (FW and backward (BW) passes of \textcolor{tritonfw}{\method \renderer} (left) and \textcolor{tritonfw}{\splatter} (right), compared to the \textcolor{pytorchfw}{Autograd renderer} and \textcolor{pytorchfw}{lifter} from ~\cite{yu2021pixelnerf,chan2023genvs}.
    \method exhibits up to 4 orders of magnitude lower memory consumption at comparable speed.
    All axes are log-scaled.
    }
    \vspace{-4mm}
    \label{fig:benchmark}
\end{figure}

%% file: sec/4_exp.tex
\section{Experiments}%
\label{sec:exp}

We first benchmark the performance of proposed components, and then demonstrate their practical usage for various 3D tasks, including single-scene optimization with image-level loss, and boosting the scalability of large-scale 3D models.

The scalability boost comes from both \emph{input-size} and \emph{modeling}.
For input-size, it dramatically increases the amount of 2D information lifted to 3D by enlarging the number of input views and the output size.
For modeling, the memory savings allow increasing the model and batch size during training.

\subsection{Memory \& speed benchmark}
We measure components' speed and memory in Figure~\ref{fig:benchmark}.
\renderer (left col.) is tested on a triplane with 256 points per ray, and compared to a PyTorch Autograd triplane renderer, adopted from~\cite{cao2023hexplane,chan2022efficient}.
It easily supports high image sizes with low memory usage, which is unaffordable for the Autograd renderer. 
\splatter (right col.) is tested on lifting $N$ input feature maps into a $160^3$ voxel grid.
We benchmark it against the lifting operations from PixelNeRF~\cite{yu2021pixelnerf} and GeNVS~\cite{chan2023genvs}, disabling the MLPs in \splatter for a fair comparison. 
As shown, \splatter can handle over a hundred views efficiently, while existing methods are restricted to just a few views. 
Speed-wise, both components are comparable to their autograd counterparts.
See supplementary material for more results.

\subsection{Single-scene Optimization with Image-level Loss.}
\input{figures/fig_single_scene.tex}
The memory efficiency of the proposed components, in particular \renderer, allows rendering high-resolution images~(\eg $512^2$) in a differentiable way with little memory overhead.
Therefore, we can seamlessly use models which take full images as input for loss calculation, \eg perceptual loss~\cite{johnson2016perceptual}, LPIPS~\cite{zhang2018perceptual}, or SDS~\cite{poole2022dreamfusion}, and backpropagate these losses back to neural 3D fields.
Constrained by memory usage, existing methods are limited to very low-resolution rendering~\cite{poole2022dreamfusion,lin2022magic3d} or complicated and inefficient deferred backpropagation~\cite{zhang2022arf}, while \method can handle image-level losses easily.
We take neural 3D field stylization as an example in Figure~\ref{fig:single_scene} and discuss more applications in supplementary.

\input{figures/fig_LRM}
\input{figures/fig_lrm_com}
\input{figures/table_LRM}

\subsection{Multi-view LRM with \method}
We first validate \method's efficacy, in particular of the triplane \renderer and \splatter, by combining \method with Large Reconstruction Model(LRM)~\cite{hong2023lrm}.
Taking four images as input, this model outputs triplane as the 3D representation via a series of transformer blocks. 
Every 3 transformer blocks (\ie 5 blocks in total), we insert the \splatter layer, which splats source view features into a new triplane, taking previous block outputs as prior shape encoding.
Plugging \method into LRM adds little computational overheads, while clearly improving the performance.
Additionally, the memory efficiency of our renderer enables LPIPS optimization without the added complexity of the deferred backpropagation in LRM \cite{hong2023lrm}.
We show the results in Table~\ref{tab:lrm} and Figure~\ref{fig:lrm},~\ref{fig:lrm_com}.

\subsection{Large-Scale 3D Reconstruction and Generation.}

\paragraph{Datasets and Baselines.}
We use CO3Dv2~\cite{reizenstein2021common} as our primary dataset, a collection of real-world videos capturing objects across 51 common categories.
We implement the versatile model for 3D reconstruction and generation as described in \cref{sec:rec}, and extend \method to unbounded scenes by contracting the ray-point's coordinates~\cite{barron2022mip} to represent background.
Without loss of generality, we utilize UNet~\cite{ronneberger2015u} with attention layers~\cite{vaswani2017attention} to process 3D hash structures.

\input{figures/fig_uncond_gen.tex}
\input{figures/table_1_dense_recon.tex}
\input{figures/table_2_uncond_gen.tex}
\paragraph{Amortized 3D Reconstruction.}
Existing amortized 3D reconstruction and novel view synthesis methods~\cite{reizenstein2021common,zhou2022sparsefusion, yu2021pixelnerf,Wang2021IBRNetLM, grf2020} only consider a few views~(up to 10) as input due to memory constraints.
Here, we enlarge the number of input views significantly.
Unlike existing category-specific models, we train a \emph{single} model on \emph{all} CO3Dv2 categories, targeting a universal reconstruction model that can work on a variety of object types, and provide useful 3D priors for the following 3D optimizations. 
During training, 20 source images from a training scene are taken as inputs and MSE losses are calculated on five other novel views.

We evaluate in two regimes:
(1) comparing our model to other feedforward baselines and single-scene overfitting methods to evaluate the model's performance;
(2) finetuning feedforward results using training views in a single scene to show the efficacy of our model as a learned 3D prior. 
Since few generalizable Nerf methods can work on all categories, 
we take ViewFormer~\cite{kulhanek22viewformer:} as the feedforward model baseline, which directly outputs novel view images using Transformer. 
In (2), we use 80 views as inputs to the feedforward model for initialization and report results of vanilla NeRF~\cite{mildenhall2020nerf}, and voxel-grid overfits (\ie, trained from scratch).
We evaluate results on novel views of unseen scenes.

Our model generates compelling reconstructions with just a single forward pass, shown in~\cref{tab:table_1_dense_recon}.
After fine-tuning, it is on par with the overfitting baselines in color accuracy (PSNR, LPIPS), but largely outperforms the hash-based baselines (Voxel) in depth error.
Since the frames of CO3D's real test scenes exhibit limited viewpoint coverage, overfitting with hashed representations leads to strong defects in geometry (see Supp.).
Here, by leveraging the memory-efficient \method for pre-training on a large dataset, our model learns a generic surface prior which facilitates defect-free geometry.
NeRF is superior in depth error while being on par in PSNR, at the cost of $\sim50\times$ longer training time.

\paragraph{Unconditional Generation.}

Our model is capable of unconditional generation with only minor modifications, specifically accepting noisy input images and rendering the clean images through a denoising process.
Utilizing the \splatter and \renderer, we can denoise multiple views (10 in experiments) within each denoising iteration, which significantly enhances the stability of the process and leads to markedly improved results.
In the inference stage, we input 10 instances of pure noise and proceed with 50 Denoising Diffusion Implicit Model (DDIM)~\cite{song2020denoising} sampling steps.
We compare our method to Viewset Diffusion~\cite{szymanowicz2023viewset} and HoloFusion~\cite{karnewar2023holofusion} quantitatively in~\cref{tab:uncond_gen} and evaluate qualitatively in \cref{fig:uncond_gen}.
Our results significantly outperform other feedforward generation models and are comparable to distillation-based method, which is very time-consuming. 

\paragraph{Conditional Generation.}
We can also introduce one clean image as conditioning, enabling single-view reconstruction.
Moreover, our framework is also amenable to extension as a text-conditioned model, utilizing captions as inputs.
We show results and comparison in Figure~\ref{fig:single_view_recon} and Supp.

%% file: figures/fig_single_scene.tex
\begin{figure}[!t]
\centering
\newcommand{\figw}{0.195\textwidth}
\begin{tabular}{ccccc}
    \includegraphics[width=\figw]{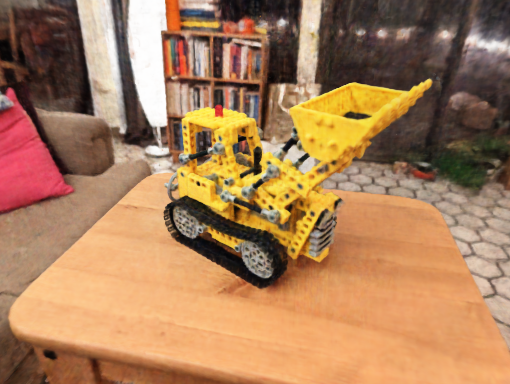}

    &\includegraphics[width=\figw]
    {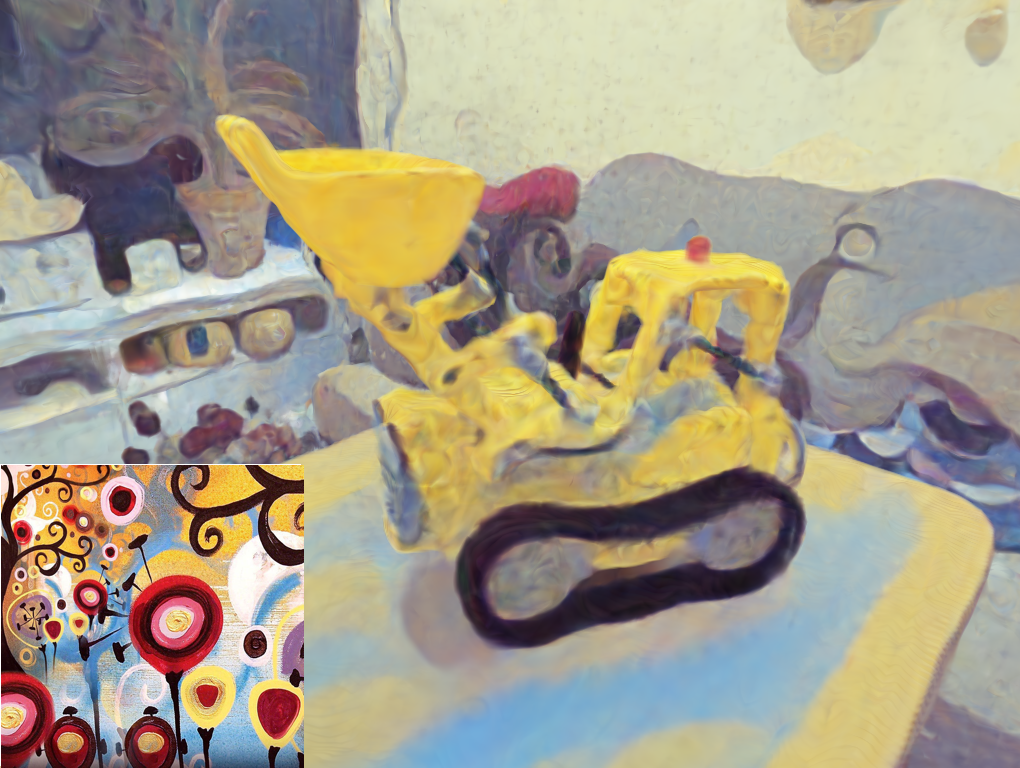}

    &\includegraphics[width=\figw]{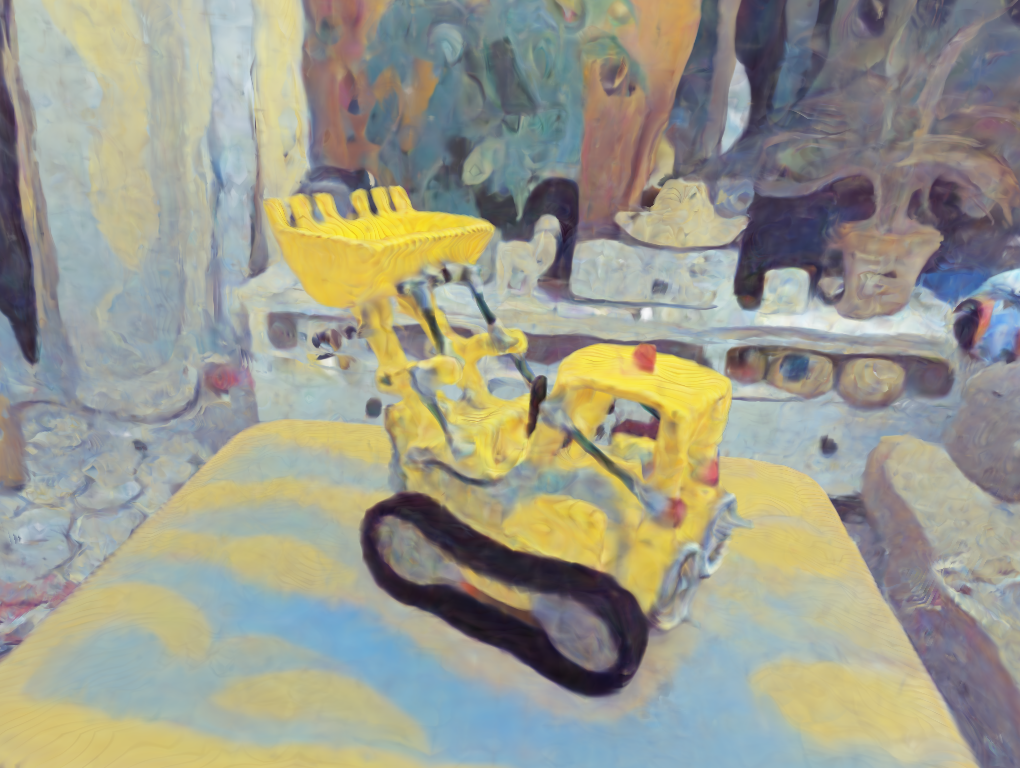}

    &\includegraphics[width=\figw]{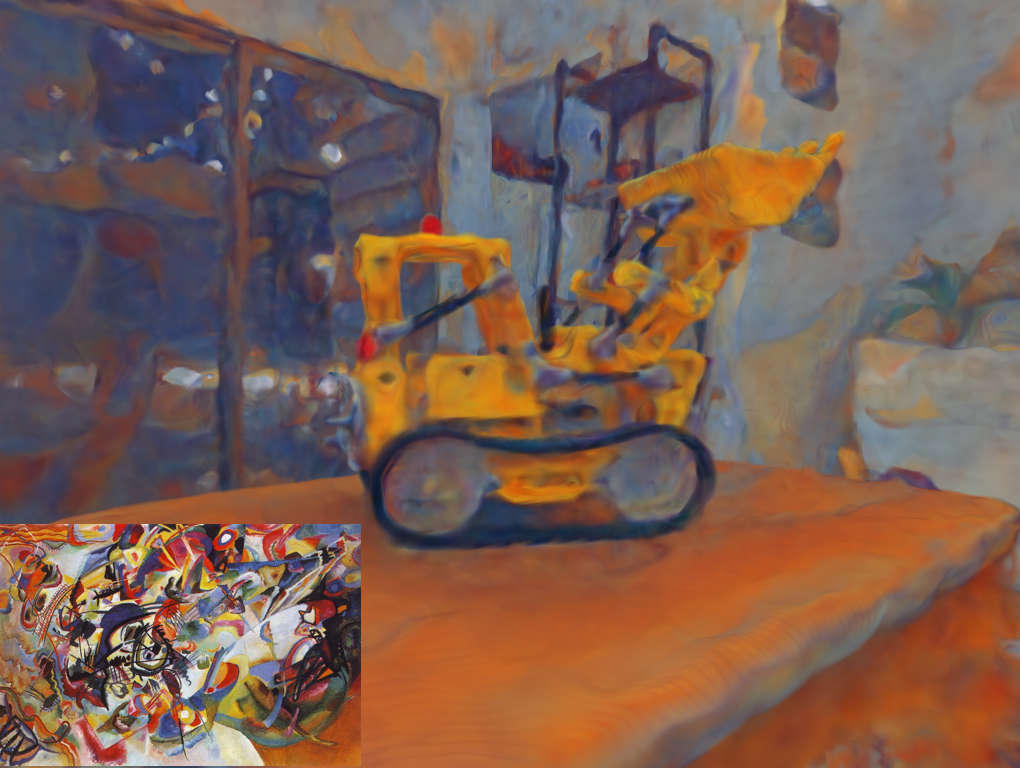}
    &\includegraphics[width=\figw]{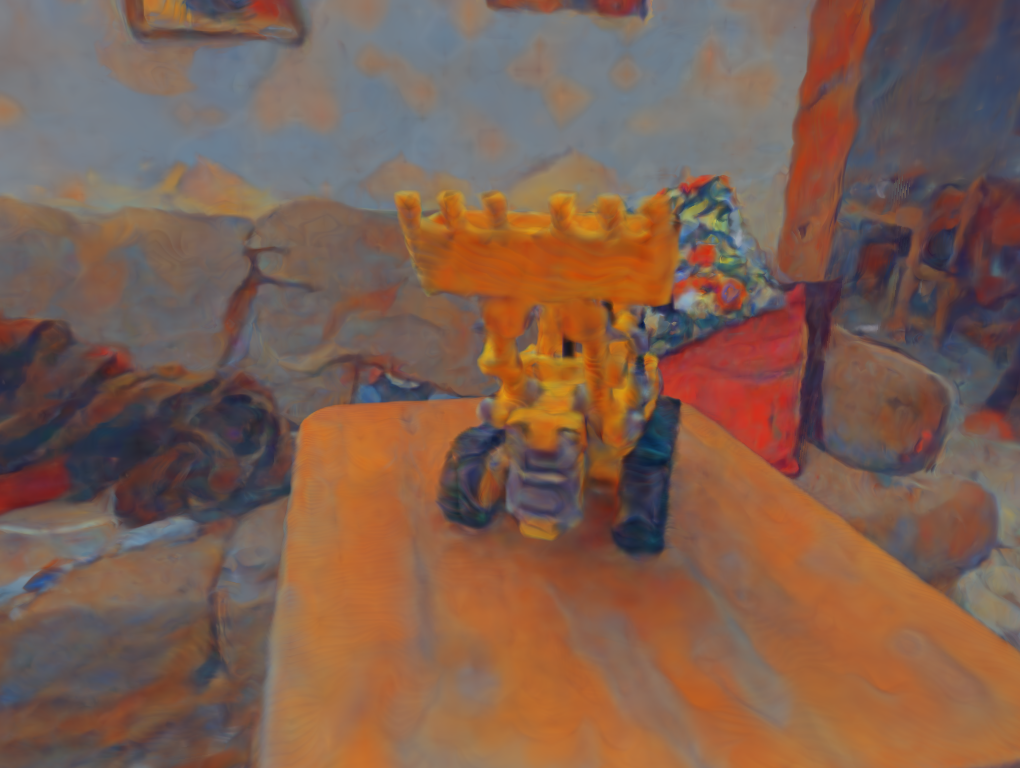} \\
    \small{Fitted 3D Scene}  &\multicolumn{2}{c}{\small Stylization 1}  &\multicolumn{2}{c}{\small Stylization 2}\\
\end{tabular}
    \vspace{-3mm}
    \caption{\textbf{Single-scene optimization with image-level losses.}
    The memory efficiency of \method allows rendering high resolution images in a differentiable way and backpropagating image-level losses.
    We show pre-optimized 3D scenes (in unseen views) and their stylizations with perceptual losses.
    }
    \label{fig:single_scene}
    \vspace{-6mm}
\end{figure}

%% file: figures/fig_LRM.tex
\begin{figure*}[t]
\centering
\newcommand{\figw}{0.16\textwidth}
\renewcommand{\arraystretch}{0}
\setlength{\tabcolsep}{0pt}
\begin{tabular}{ccccccc}
\includegraphics[width=\figw]{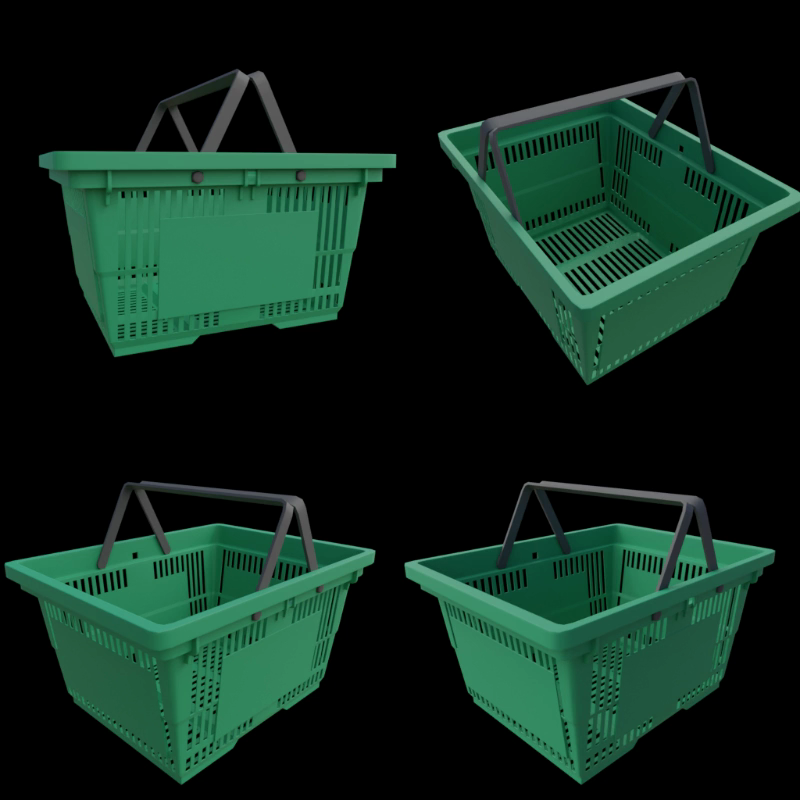}&
\includegraphics[width=\figw]{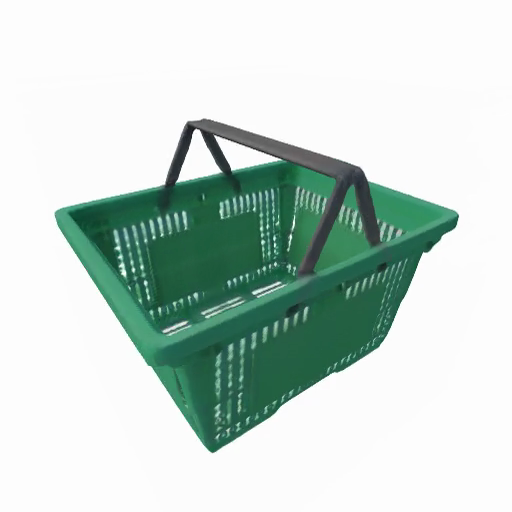}&
\includegraphics[width=\figw]{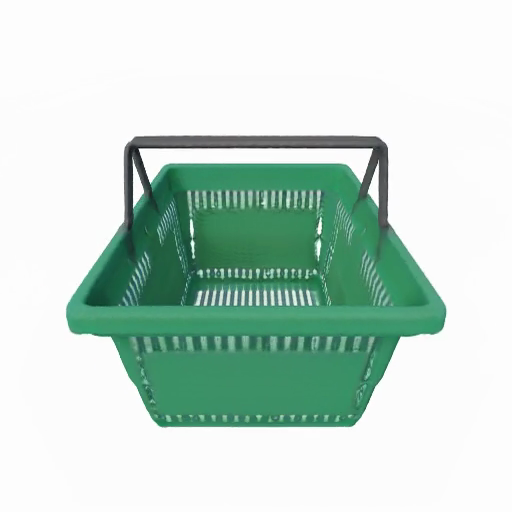}&
\includegraphics[width=\figw]{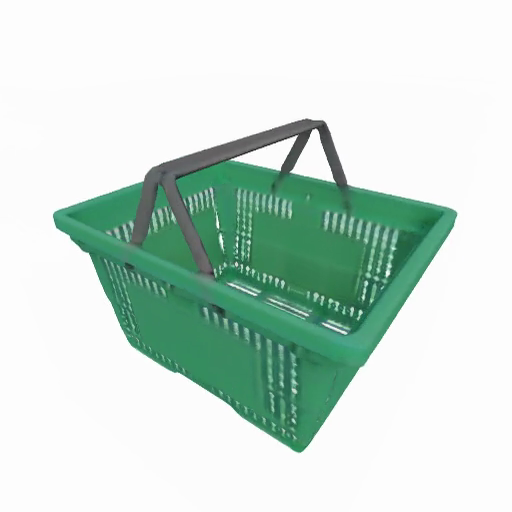}&
\includegraphics[width=\figw]{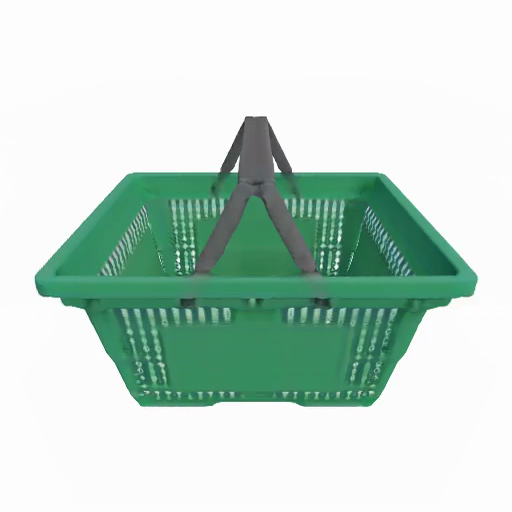}&
\includegraphics[width=\figw]{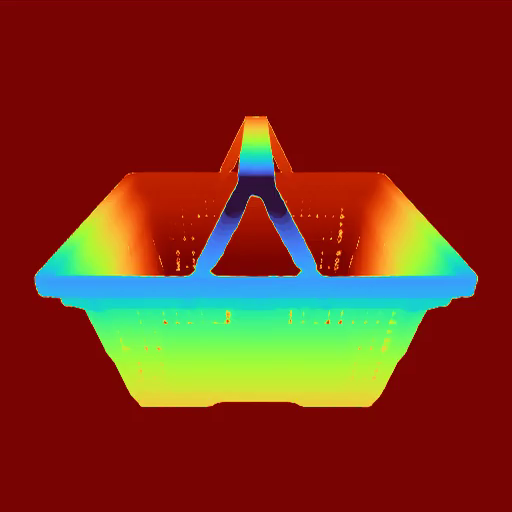}\\
\includegraphics[width=\figw]{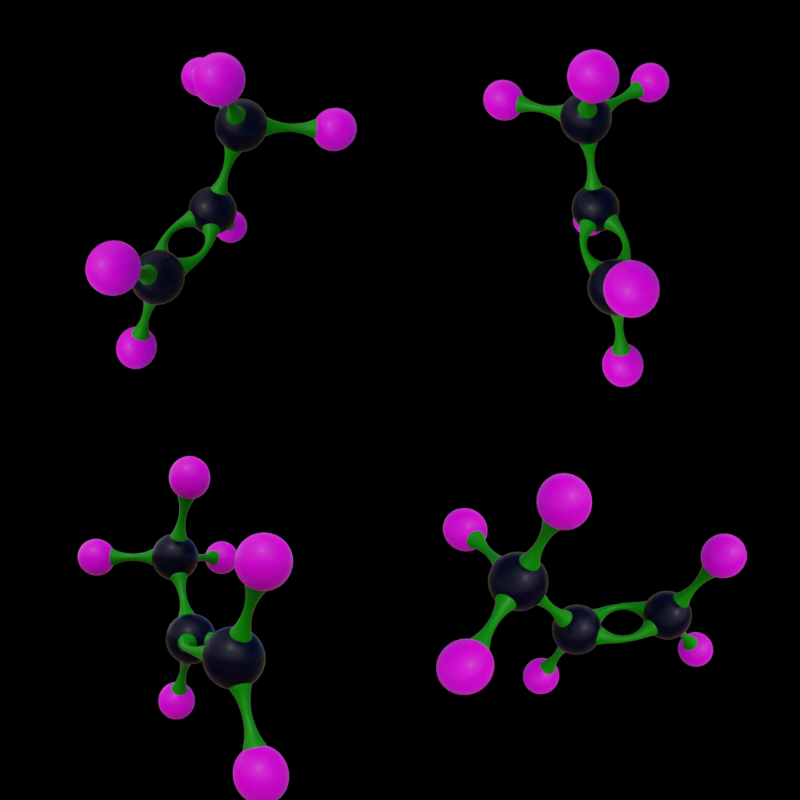}&
\includegraphics[width=\figw]{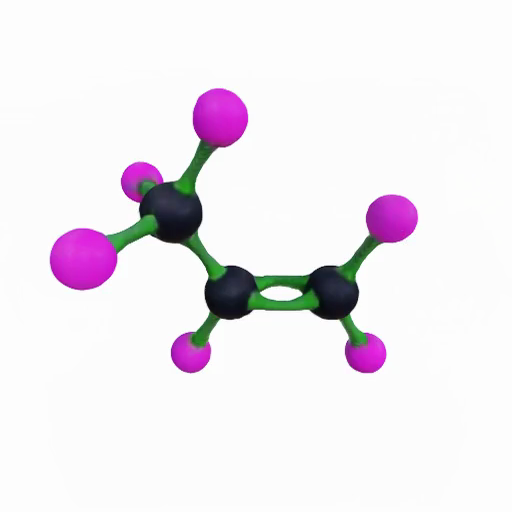}&
\includegraphics[width=\figw]{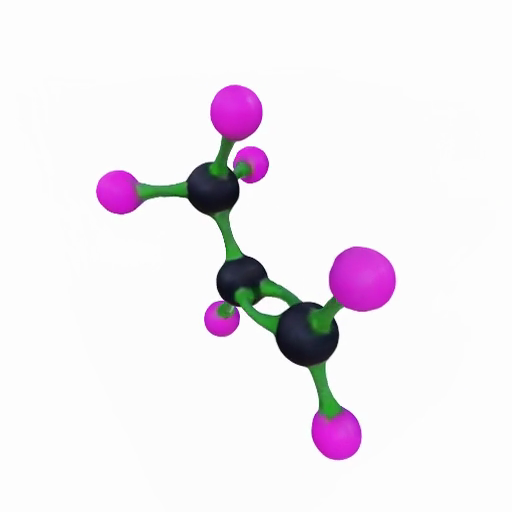}&
\includegraphics[width=\figw]{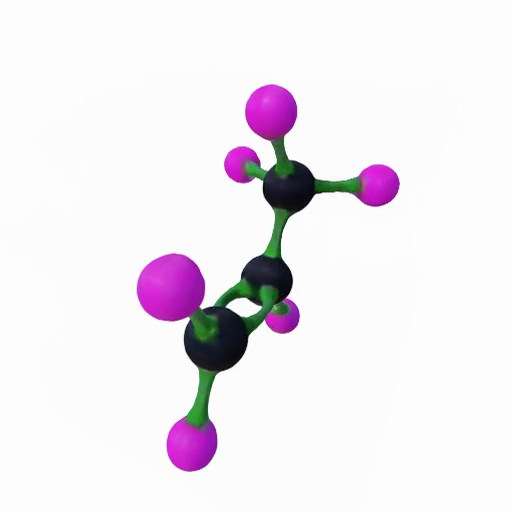}&
\includegraphics[width=\figw]{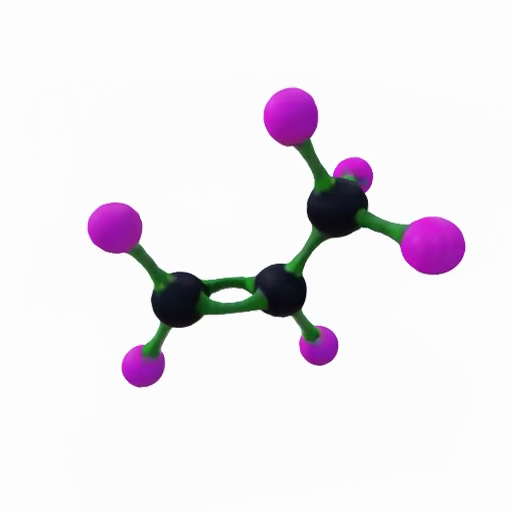}&
\includegraphics[width=\figw]{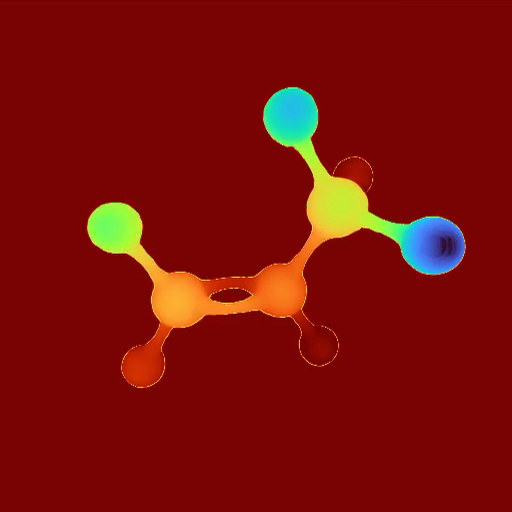}
\end{tabular}
\vspace{-3mm}
\caption{%
\textbf{Multi-view Large Reconstruction Model (LRM) with \method.}
Taking four views as input (leftmost column), we show the RGB renders (mid) and depth (rightmost column) of the 3D reconstruction.%
}
\label{fig:lrm}%
\vspace{-4mm}
\end{figure*}

%% file: figures/fig_lrm_com.tex
\begin{figure*}[t]
\centering
\includegraphics[width=1.0\textwidth]{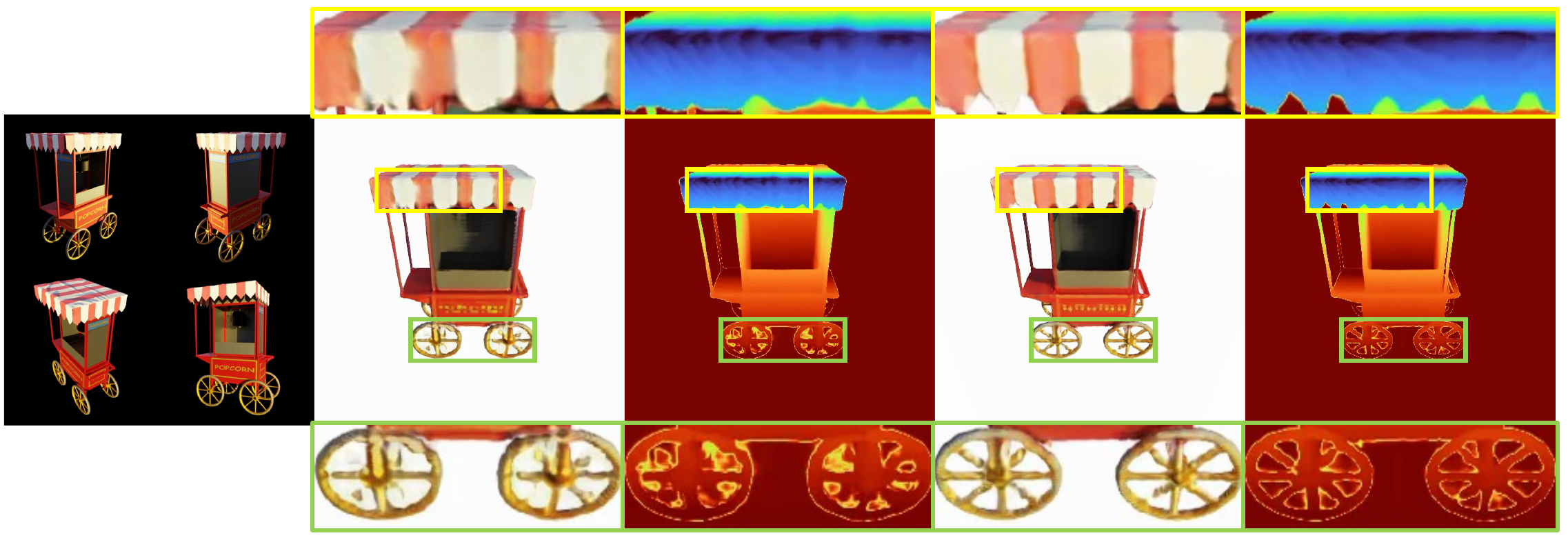} 
\setlength\tabcolsep{0pt}
\newlength\q
\setlength\q{\dimexpr .3\textwidth -2\tabcolsep}
\noindent\begin{tabular}{p{\q}p{\q}p{\q}} %
\text{Input} & \text{LRM} & \text{LRM + \method} \\
\end{tabular}
\vspace{-2mm}
\caption{%
\textbf{Visual Comparison of LRM}.
Adding \method to LRM gives more accurate geometry and appearance with little additional computation and memory cost.%
}
\label{fig:lrm_com}
\vspace{-3mm}
\end{figure*}

%% file: figures/table_LRM.tex
\vspace{0.2cm}%
\noindent\begin{minipage}[t]{0.48\textwidth}%
\centering%
\vspace{-1.6cm}%
\captionof{table}{%
  \textbf{Quantitative results of LRM.} The proposed method could effectively improve the reconstruction results, especially geometry (depth L1). }
    \resizebox{\textwidth}{!}{%
  \begin{tabular}[b]{lcccc}
    \toprule
        Method &  PSNR$\uparrow$ & LPIPS$\downarrow$ & IOU$\uparrow$& Depth L1$\downarrow$\\
        \midrule
        LRM~\cite{hong2023lrm} &23.7 & 0.113 & 0.904 & 0.208  \\
       \method+LRM & \textbf{24.1} & \textbf{0.106} & \textbf{0.916} & \textbf{0.168} \\
       \bottomrule
  \end{tabular}
  } %
  \label{tab:lrm}
\end{minipage}%
\hfill
\begin{minipage}[t]{0.48\textwidth}
\centering
\resizebox{\textwidth}{!}{
    \begin{tabular}[t]{cccc}
    \includegraphics[height=0.125\textwidth]{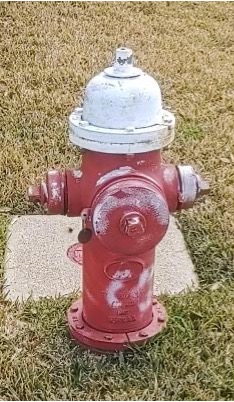} &\includegraphics[width=0.125\textwidth]{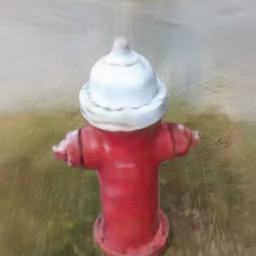}
    &\includegraphics[width=0.125\textwidth]{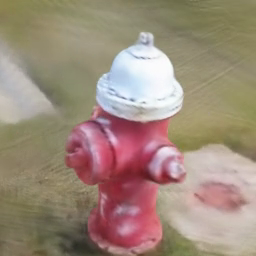}  &\includegraphics[width=0.125\textwidth]{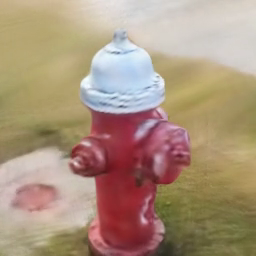} 
        \end{tabular}
}%
    \vspace{-5mm}
    \captionof{figure}{
  \textbf{Monocular 3D Reconstruction.}
  With a single clean image as input (1st col.), our model could generate realistic 3D structures matching the input.}
  \label{tab:LRM}
\end{minipage}
\vspace{-4mm}

%% file: figures/fig_uncond_gen.tex
\begin{figure*}[t]
\centering
\newcommand{\figw}{0.16\textwidth}
\renewcommand{\arraystretch}{0}
\setlength{\tabcolsep}{0pt}
\begin{tabular}{ccccccc}
\includegraphics[width=\figw]{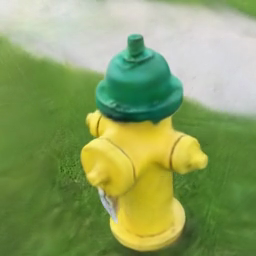}&
\includegraphics[width=\figw]{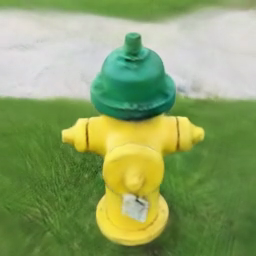}&
\includegraphics[width=\figw]{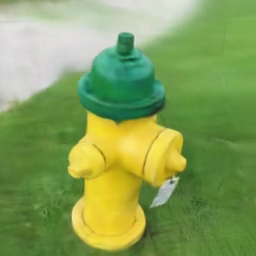}&
\includegraphics[width=\figw]{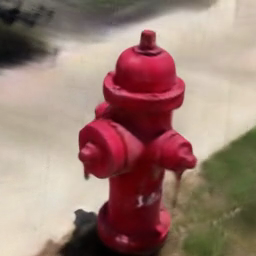}&
\includegraphics[width=\figw]{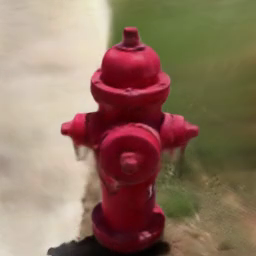}&
\includegraphics[width=\figw]{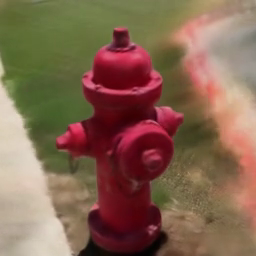}\\
\includegraphics[width=\figw]{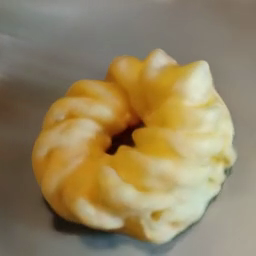}&
\includegraphics[width=\figw]{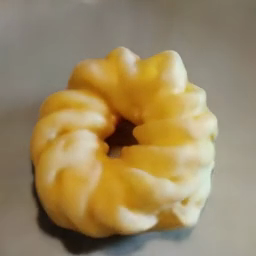}&
\includegraphics[width=\figw]{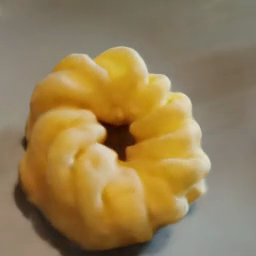}&
\includegraphics[width=\figw]{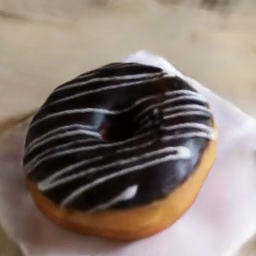}&
\includegraphics[width=\figw]{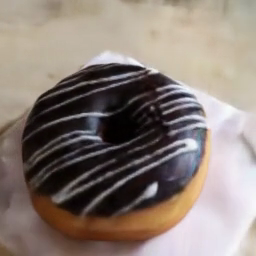}&
\includegraphics[width=\figw]{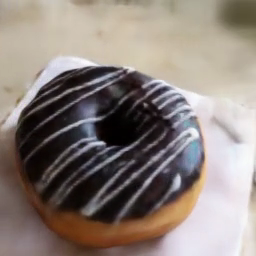}
\end{tabular}
\vspace{-2mm}
\caption{%
\textbf{Unconditional 3D Generation}
displaying samples from our \method-augmented Viewset Diffusion trained on CO3Dv2~\cite{reizenstein2021common}.%
}
\label{fig:uncond_gen}%
\vspace{-3mm}
\end{figure*}

%% file: figures/table_1_dense_recon.tex
\begin{table}[!t]
    \centering
    \newcommand{\bof}[1]{\textbf{#1}}
    \newcommand{\ul}[1]{\underline{#1}}
    \caption{%
    \textbf{Amortized 3D Reconstruction.}
    Our feedforward reconstructor~(\method) trained on the \emph{whole} CO3Dv2 significantly outperforms  baseline ViewFormer~\cite{Kulhanek2022ViewFormerNN}.
    We further compare overfitting baselines (Voxel, NeRF~\cite{mildenhall2020nerf}) to \method, and to their scene-tuned versions (``Feedforward + Overfit'').
    Initializing from \method-feedforward removes defective geometry leading to better depth error.}%
    \vspace{-2mm}
    \resizebox{\columnwidth}{!}{%
    \setlength{\tabcolsep}{3pt}
    \begin{tabular}{llrrrrr} \toprule
    Method         & Mode & \#views & PSNR$\uparrow$ & LPIPS$\downarrow$ & Depth corr.$\uparrow$ & Time$\downarrow$  \\ \midrule
    ViewFormer~\cite{Kulhanek2022ViewFormerNN} & Feedforward & 9 & 16.4 & 0.274 &N/A  &N/A\\
    \method    & Feedforward & 10  & 20.7 & 0.141 & 0.356 & 1.6 sec \\
    \method   & Feedforward & 20  & 20.9 & 0.136 & 0.382 & 1.9 sec \\
    \method   & Feedforward & 40  & 21.4 & \ul{0.131} & \ul{0.405} & 2.5 sec \\
    \rowcolor[HTML]{EDF3F9}
    \method    & Feedforward + Overfit    & 160 & \ul{26.2} & \bof{0.086} & \bof{0.449} & 5 min   \\
    \rowcolor[HTML]{EDF3F9}
    Voxel         & Overfit from scratch    & 160 & \bof{26.5} & \bof{0.086} & 0.373 & 35 min  \\ \midrule
    \rowcolor[HTML]{EDF3F9}
    NeRF          & Overfit from scratch    & 160 & 26.3 & 0.108 & 0.658 & 1 day  \\ \bottomrule
    \end{tabular}
    }%
    \label{tab:table_1_dense_recon}
    \vspace{-4mm}
    \end{table}

%% file: figures/table_2_uncond_gen.tex
\begin{table}[t]
\caption{\textbf{Unconditional 3D Generation on CO3Dv2.}
Our \method significantly outperforms HoloDiffusion~\cite{karnewar2023holodiffusion} and Viewset Diffusion~\cite{szymanowicz2023viewset}.
It even beats HoloFusion~\cite{karnewar2023holofusion}, a distillation-based method, which takes 30 mins for one generation.}
\vspace{-2mm}
\resizebox{\linewidth}{!}{
\small
\newcommand{\ul}[1]{\underline{#1}}
\setlength{\tabcolsep}{1pt}
\begin{tabular}{lcrrrrrrrrrrc}
\toprule
Method & Feed- & 
\multicolumn2c{\texttt{Hydrant}} &
\multicolumn2c{\texttt{Teddybear}} &
\multicolumn2c{\texttt{Apple}} &
\multicolumn2c{\texttt{Donut}} &
\multicolumn2c{Mean} &
Inference
\\
\cmidrule(lr){3-4} \cmidrule(lr){5-6} \cmidrule(lr){7-8} \cmidrule(lr){9-10}
\cmidrule(lr){11-12}
&forward
& 
\multicolumn1c{\footnotesize{FID $\downarrow$}} & \multicolumn1c{\footnotesize{KID $\downarrow$}} &
\multicolumn1c{\footnotesize{FID $\downarrow$}} & \multicolumn1c{\footnotesize{KID $\downarrow$}} &
\multicolumn1c{\footnotesize{FID $\downarrow$}} & \multicolumn1c{\footnotesize{KID $\downarrow$}} &
\multicolumn1c{\footnotesize{FID $\downarrow$}} & \multicolumn1c{\footnotesize{KID $\downarrow$}} &
\multicolumn1c{\footnotesize{FID $\downarrow$}} & \multicolumn1c{\footnotesize{KID $\downarrow$}} &
Time \\
\midrule
\rowcolor[HTML]{EDF3F9}
HoloFusion~\cite{karnewar2023holofusion} & $\times$ &\textbf{66.8} & \textbf{0.047} & \textbf{87.6} & 0.075 &\ul{69.2} &\ul{0.063} &\ul{109.7} &\ul{0.098} &\ul{83.3} &\ul{0.071} & 30mins\\
HoloDiffusion~\cite{karnewar2023holodiffusion} & \checkmark &100.5 &0.079 &109.2 &0.106 &94.5 &0.095 &115.4  & 0.085 &122.5 & 0.102 &<2min\\
Viewset Diffusion~\cite{szymanowicz2023viewset} &\checkmark &150.5 &0.124 &219.7 &0.178 &- &- & - &- & - &- &<2min\\
\method &\checkmark &\ul{75.1} &\ul{0.058} &87.9 &\textbf{0.070} &\textbf{32.6} &\textbf{0.019} &\textbf{44.0} &\textbf{0.019} &\textbf{59.9} &\textbf{0.042} &<2min \\
\bottomrule
\end{tabular}} %
\label{tab:uncond_gen}%
\vspace{-6mm}
\end{table}

%% file: sec/5_conclusion.tex
\section{Conclusion}%
\label{sec:conclusion}

We have introduced \method, a versatile framework that provide two  novel components, \splatter and \renderer, which address the key memory bottleneck in network that manipulate neural fields.
We have showcased the potential of these primitives in a number of applications, boosting models for reconstruction, generation and more.
Once released to the community, we hope that these primitives will be used by many to boost their own research as well.
\footnote{We discuss limitations and potential negative impact in Supp.}

\section{Acknowledgement}%
\label{sec:acknowledgement}
This work was done during Ang Cao's internship at Meta AI as well as at the University of Michigan, partially supported by a grant from LG AI Research.
We thank Roman Shapovalov, Jianyuan Wang, and Mohamed El Banani for their valuable help and discussions.

%% file: sec/supp_text.tex
\newcommand{\norm}[1]{\left\lVert#1\right\rVert}

\section{Supplementary Videos}
Please watch our attached video for a brief summary of the paper and more results.
We include more generated results and 360-degree rendering videos from our reconstructed and generated 3D structures to show their 3D consistency. 

\section{Social Impact}
Our main contribution is \method \splatter and \renderer,  a pair of 3D components which could be used to significantly scale the mapping between 2D images and neural 3D fields. 
Beyond their integral role in our versatile pipeline for 3D reconstruction and generation, single scene optimization, and LRM with \method, these components can also function as highly scalable plug-ins for various 3D applications. We earnestly hope that they will be instrumental in advancing future research.

Based on \method \splatter and \renderer, we have established a comprehensive framework for 3D reconstruction and generation. 
Similar to many other generative models~\cite{szymanowicz2023viewset, chan2023genvs, liu2023zero}, it is important to note that the results generated by this framework have the potential to be used in the creation of synthetic media.

\section{Limitations \& Discussions}
Our motivation of introducing contract coordinates~\cite{barron2022mip} is to assist the model in differentiating between foreground and background elements, thereby enhancing the quality of foreground generation and reconstruction.
Although contract coordinates could represent unbounded scenes,
our main focus is still on foreground objects, 
and reconstructing or generating unbounded backgrounds is beyond the scope of this paper.
Therefore, we only sample limited points in unbounded regions, which leads to floaters, blurriness and clear artifacts in the background, as can be observed in videos. 
Also, generating diverse and realistic backgrounds is a challenging task and we leave it as a promising future direction.

\method introduces a versatile approach for scaling the mapping between 2D and 3D in neural 3D fields, designed to be compatible with arbitrary 3D hash representations with differentiable sampling functions.
While our validation of this design has focused on voxel and triplane models, its adaptability should allow for easy generalization to other 3D hash representations, such as Hash Table~\cite{instant3d2023} or HexPlane~\cite{cao2023hexplane,fridovich2023k}.
We pick voxel grids and triplanes as their structures are easy to be processed by the existing neural networks while designing neural networks to process some other 3D hash structures like hash tables is still an open question. 
Developing neural networks to support other 3D hash structures is a promising direction to explore while beyond the scope of this paper.

 \method significantly solves the memory bottlenecks in neural 3D fields, making rendering and splatting a large number of images possible in the current 3D pipelines. 
 Although \method has comparable speed to existing methods, rendering and splatting a large number of images is still time-consuming, which may limit its utilization in real applications. 
 For example, doing a forward and backward pass on $512 \times 512$ rendered images takes around 5 seconds for each iteration.
 For \renderer, the spent time grows linearly to the ray numbers when ray numbers are huge. 
 Reducing the required time for large ray numbers would be a promising direction.

Sadly, we observe a performance gap between different 3D hash representations (\ie, voxel grids and triplanes) in the versatile 3D reconstruction and generation framework. 
Without loss of generalization, we use 3D UNet to process voxel grids and 2D UNet to process Triplane.  
Three planes~(XY, YZ, ZX) are concatenated into a single wide feature map and fed to 2DUNet. 
The self-attention mechanism is then applied across all patches from the three planes, making this network an extension of our 3DUNet designed for voxel grids.
However, we observed that this neural network configuration does not yield flawless results. In 3D reconstruction tasks, the images rendered at novel viewpoints exhibit slight misalignments with the ground-truth images. For generative tasks, while the network can produce realistic samples, it occasionally generates flawed outputs that significantly impact the Fidelity (FID) and Kernel Inception Distance (KID) scores. Developing a more efficacious neural network model for TriPlane processing~\cite{wang2022rodin, instant3d2023, cao2023large}, which could effectively communicate features from three planes, presents a promising avenue for future research.

\section{\method Details}

\subsection{Implementation Details}
\paragraph{Normalization Process in \splatter.}

Starting from a zero-initialized hash $ \theta$, \splatter is done by accumulating $\bv_{ij}$ to the hash cell (\ie voxel grids or triplanes) that contain $\x_{ij}$, using the same trilinear/bilinear weights used in the \renderer operator to sample $\theta$.
After accumulating over all $M$ rays, each hash cell is normalized by the sum of all splatting bi/trilinear weights landing in the cell.
The normalization operation employed in our method, analogous to average pooling, averages the information splatted at identical positions in the hash $\theta$.
This process guarantees that the magnitudes of the splatted features are comparable to those of the input view features, a factor that is beneficial for the learning process.

In the actual implementation, we execute the splatting process twice within the \splatter kernel. 
Initially, we splat the features of the input image into $\theta$.
 Subsequently, a second set of weight maps is created, matching the spatial dimensions of the input image features, but with a feature  of a single-scale: 1
These weight maps are then splatted into $\theta_{weight}$.
During the second splatting process within the \splatter kernel, we deactivate the Multilayer Perceptrons (MLPs) and suspend sampling from prior hash representations. 
This modification is implemented because our objective is to tally the frequency and weights of points being splatted into the same position within the hash representations,
instead of learning to regress features. 
Finally, we get $\theta / \theta_{weight}$. 

Performing the splatting operation twice inevitably results in additional time and memory overhead.
In practice, $\theta_{weight}$ is relatively lightweight while $\theta$ is more memory-intensive.
This is because 
they have the same spatial shape while $\theta_{weight}$ has a feature dimension of only 1. 
The normalization step $ \theta / \theta_{weight}$, which is implemented in PyTorch, will cache the heavy $\theta $, thereby increasing memory usage. 
We manually cache $\theta_{weight}$ to normalize gradients during backpropagation.

\paragraph{Experimental Details.}
We use $160\times 160 \times 160$ voxel grids and $160\times 160$ triplanes in our model.
The input images are processed using a VAE-encoder~\cite{rombach2021highresolution} trained on the ImageNet dataset~\cite{deng2009imagenet} and are converted into 32-dimensional feature vectors. 
Both the \splatter and \renderer components are equipped with 3-layer MLPs with a width of 64. Regarding training, we conduct 1000 iterations per epoch. The generative model is trained over 100 epochs, taking approximately 4 days, while the reconstruction model undergoes 150 epochs of training, lasting around 6 days, on a setup of 16 A100 GPUs, processing the entire Co3Dv2 dataset. 

For \splatter, we sample 160 points along the ray. 
For \renderer, we sample 384 points along the ray, rendering $256 \times 256$ images. 
Instead of using original contract coordinates~\cite{barron2022mip}, we use a slightly different version which maps unbounded scenes into a  $[-1, 1]$ cube.
\begin{equation}
\operatorname{CC}(\mathbf{x})  = 0.5 * \begin{cases}
a  * \mathbf{x} & \norm{\mathbf{x}} \leq 1\\
\left((2-a)* (1  - \frac{1}{\norm{\mathbf{x}}}) + a \right)\left(\frac{\mathbf{x}}{\norm{\mathbf{x}}}\right) & \norm{\mathbf{x}} > 1 \label{eq:contract}
\end{cases}
\end{equation}

We introduce a scale $a$ to control the ratio between foreground and background regions, where the foreground regions are mapped to $[-a/2, a/2]$.
As we are using explicit 3D hash, mapping foreground regions into larger regions would be helpful to represent details. 
When $a=1$, it becomes the normal contract coordinates. 
We convert $X, Y, Z$ axes into contract coordinates independently.

\subsection{\method Performance Benchmark}
\input{figures/fig_renderer_benchmark_all}

Besides Autograd Renderer, implemented by pure Pytorch, we additionally compare  \method \renderer to two baselines: \emph{Checkpointing} and \emph{NerfAcc's Instant-NGP}, shown in Figure~\ref{fig:supp_renderer_bench}.

\emph{Checkpointing} baseline applies the checkpointing technique in Pytorch to Atugograd Renderer, which na\i ve recalculates forward pass results during backward pass to save memories. 
Trivially applying checkpointing on Autograd indeed saves memories both in forward pass and backward pass, while still requires a large amount of memories, and cannot be used for large ray numbers.

\emph{NerfAcc's Instant-NGP} is the Instant-NGP~\cite{muller22instant} implemented by NerfAcc~\cite{li22nerfacc:}, which is claimed to be 1.1$\times$ faster than the original version of Instant-NGP, with tremendous optimization tricks for speed. 
Instant-NGP combines hash grid as 3D structures with fused MLP kernels (tiny-cuda-dnn), which is different from our \renderer with triplanes as 3D structures, and its internal settings are less flexible to change. 
To this end, it is hard to do a perfectly fair comparison. 
But still, we found that instant-NGP cannot work (will crash) with large image sizes, as they heavily rely on the L2 cache of GPUs for optimal speed, which memory is very limited and cannot support large image sizes. 
While their backward pass speed is significantly faster than \method \renderer, it still cannot be extended to large output image sizes. 

\section{More Results}

\subsection{Single Scene Optimization}

\input{figures/table_supp_nerf}
\paragraph{Synthetic NeRF Results.}
We validate the correctness of \method by overfitting on the Synthetic NeRF dataset, shown in Table~\ref{tab:supp_nerf}.
As the target is to show the convergence of \method, we don't employ any complicated tricks to optimize the performance and speed.
\method could get promising single-scene optimization results, demonstrating that it could be used as a reliable package in various 3D tasks.

\paragraph{DreamFusion with SDS Loss.}
The memory efficiency of \method allows directly applying SDS~\cite{poole22dreamfusion:} on high-resolution rendered images. 
As analyzed in Magic3D~\cite{lin22magic3d:}, existing 3D generations using SDS loss are limited to low-resolution rendered images: they first render low-resolution images for SDS to generate coarse 3D structures, and then convert the generated 3D structures into 3D meshes, which are used to generate high-resolution images. 
Using \method allows direct optimization on high-resolution images. 

\input{figures/fig_adverserial}
\paragraph{Adversarial Attacking on LVM~(Large Vision Model).}
We showcase another interesting application empowered by our \method by adversarial attacking LVM models, \eg CLIP~\cite{radford2021learning} and BLIP2~\cite{li2023blip}
After rendering full images from the  neural 3D field overfitted on a specific scene, we feed rendered images into CLIP model and calculate cosine similarity between image feature vectors and target text vectors, which similarity works as a loss to optimize the neural 3D fields.

\subsection{Multi-view LRM with \method}
We show more results of Multi-view LRM with \method in Figure~\ref{fig:supp_lrm_com} and Figure~\ref{fig:sup_lrm}.
\input{figures/fig_sup_lrm_comparison}
\input{figures/fig_sup_LRM}

\subsection{3D Reconstruction}
We show amortized 3D reconstruction results after fine-tuning on a single scene in Figure~\ref{fig:3D_recon}, with voxel grids (\method-\texttt{Vox}) and triplanes (\method-\texttt{Tri}) as 3D structures.
We compare them to overfitting results (training from scratch) using the 3D structures. 
Overfiting a single scene on Co3Dv2 dataset leads to defective 3D structures, like holes in depths. 
Initializing from the outputs of our amortized 3D reconstruction model could effectively solve this problem, leading to better results. 
\input{figures/fig_recon_depth}

\subsection{Unconditional Generation}
We show 360-degree rendering for unconditional generation in Figure~\ref{fig:sup_uncond_gen_1} and Figure~\ref{fig:sup_uncond_gen_2}.
\input{figures/fig_sup_uncond}
\input{figures/fig_sup_uncond_2}

\subsection{Conditioned Generation}
We show monocular 3D reconstruction with a single image as input in Figure~\ref{fig:single_view_recon}, and text-conditioned generation in Figure~\ref{fig:text_3D_co3d}.
For text-conditioning experiments, we follow CAP3D~\cite{luo2023scalable}: we use BLIP2~\cite{li2023blip} to generate captions of each image insides scenes and utilize LLAMA2 to output the comprehensive caption for the whole scene. 
\input{figures/fig_cond_gen}
\input{figures/fig_text_3D_co3d}

%% file: figures/fig_renderer_benchmark_all.tex
\begin{figure}[!t]
    \centering
    \definecolor{lightplane}{RGB}{137, 207, 240}
    \definecolor{naive}{RGB}{96, 130, 182}
    \definecolor{checkpoint}{RGB}{255, 208, 111}
    \definecolor{nerfaccl}{RGB}{231, 98, 84}

    \includegraphics[width=1.0\textwidth]{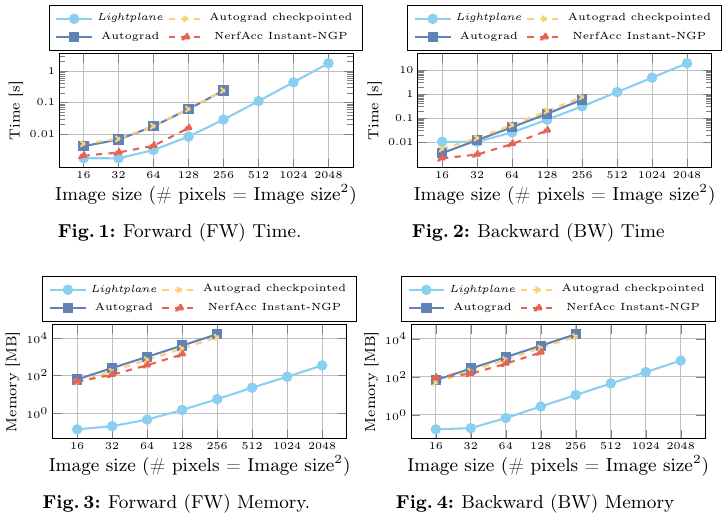}
    \caption{\textbf{\method \renderer memory \& speed benchmark} showing the forward (FW and backward (BW) passes of \textcolor{lightplane}{\method \renderer}, compared to the \textcolor{naive}{Autograd renderer}, \textcolor{checkpoint}{Checkpointing}~(Pytorch checkpointing on \textcolor{naive}{Autograd renderer}), and \textcolor{nerfaccl}{NerfAcc Instant-NGP}~(Instant-NGP~\cite{muller2022instant}  implemented in NerfAcc~\cite{li2023nerfacc}, which is claimed 1.1$\times$ faster than the original version) 
    \method exhibits up to 4 orders of magnitude lower memory consumption at comparable speed.
    All axes are log-scaled.
    }
    \label{fig:supp_renderer_bench}
\end{figure}

%% file: figures/table_supp_nerf.tex
\begin{table}[!t]
    \centering
    \newcommand{\bof}[1]{\textbf{#1}}
    \newcommand{\ul}[1]{\underline{#1}}
    \caption{%
    \textbf{Quantitative results on NeRF Synthetic dataset~\cite{mildenhall20nerf:}.}}
    \setlength{\tabcolsep}{5pt}
    \begin{tabular}{l|ccc}
    \toprule
        Method & PSNR$\uparrow$ & SSIM$\uparrow$
        & $\text{LPIPS}_{\text{VGG}}\downarrow$\\
        \midrule
         NeRF~\cite{mildenhall2020nerf} &31.01 &0.947 &0.081 \\
         Plenoxels~\cite{yu2021plenoxels} &31.71 &0.958 &\bof{0.049} \\
         DVGO~\cite{Sun2021DirectVG} &31.95 &0.957 &0.053 \\
         TensoRF-CP-384~\cite{chen22tensorf:} &31.56 &0.949 &0.076 \\
         TensoRF-VM-48~\cite{chen22tensorf:} &\bof{32.39} &\bof{0.957} &0.057 \\
         \method &\ul{32.12} &\bof{0.957} & \ul{0.050}\\ 
        \bottomrule
    \end{tabular}
    \label{tab:supp_nerf}
    \end{table}

%% file: figures/fig_adverserial.tex
\begin{figure}
\centering
\setlength\tabcolsep{1pt}
\begin{tabular}{cccc}

\includegraphics[width=0.24\textwidth]{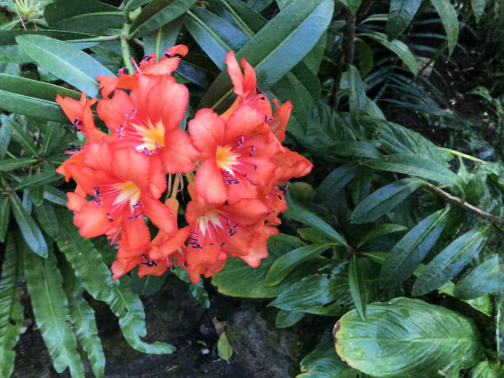}  &\includegraphics[width=0.24\textwidth]{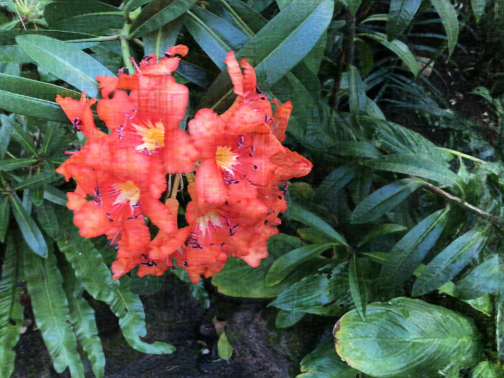}
&\includegraphics[width=0.24\textwidth]{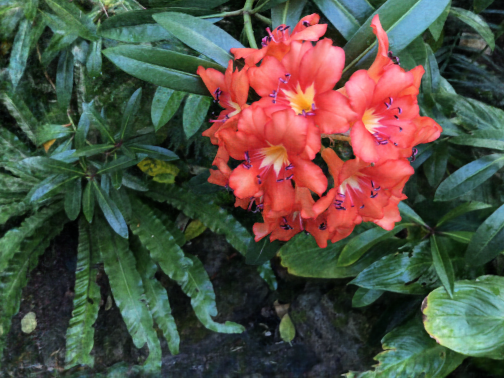}  &\includegraphics[width=0.24\textwidth]{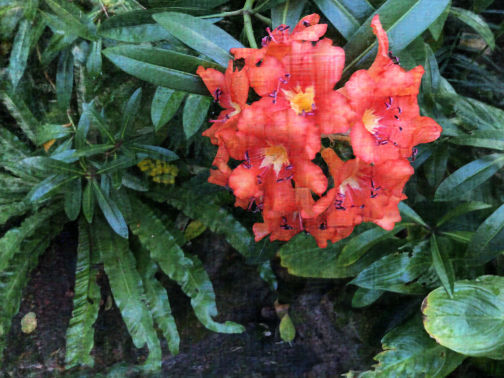}\\
\text{Fitted}
&\text{Attacked}
&\text{Fitted}
&\text{Attacked}
\end{tabular}%
\caption{
    \textbf{3D Adversarial Attacking on CLIP model.}
    Given a fitted 3D scene (1st and 3rd column), we optimize the neural 3D fields so that features of rendered images are aligned to a specific text description, \ie giraffe, in CLIP's feature space,  while keeping the appearance perceptually the same. 
}%
\label{fig:sup_attack}%
\end{figure}

%% file: figures/fig_sup_lrm_comparison.tex
\begin{figure}
    \centering
    \includegraphics[width=\textwidth]{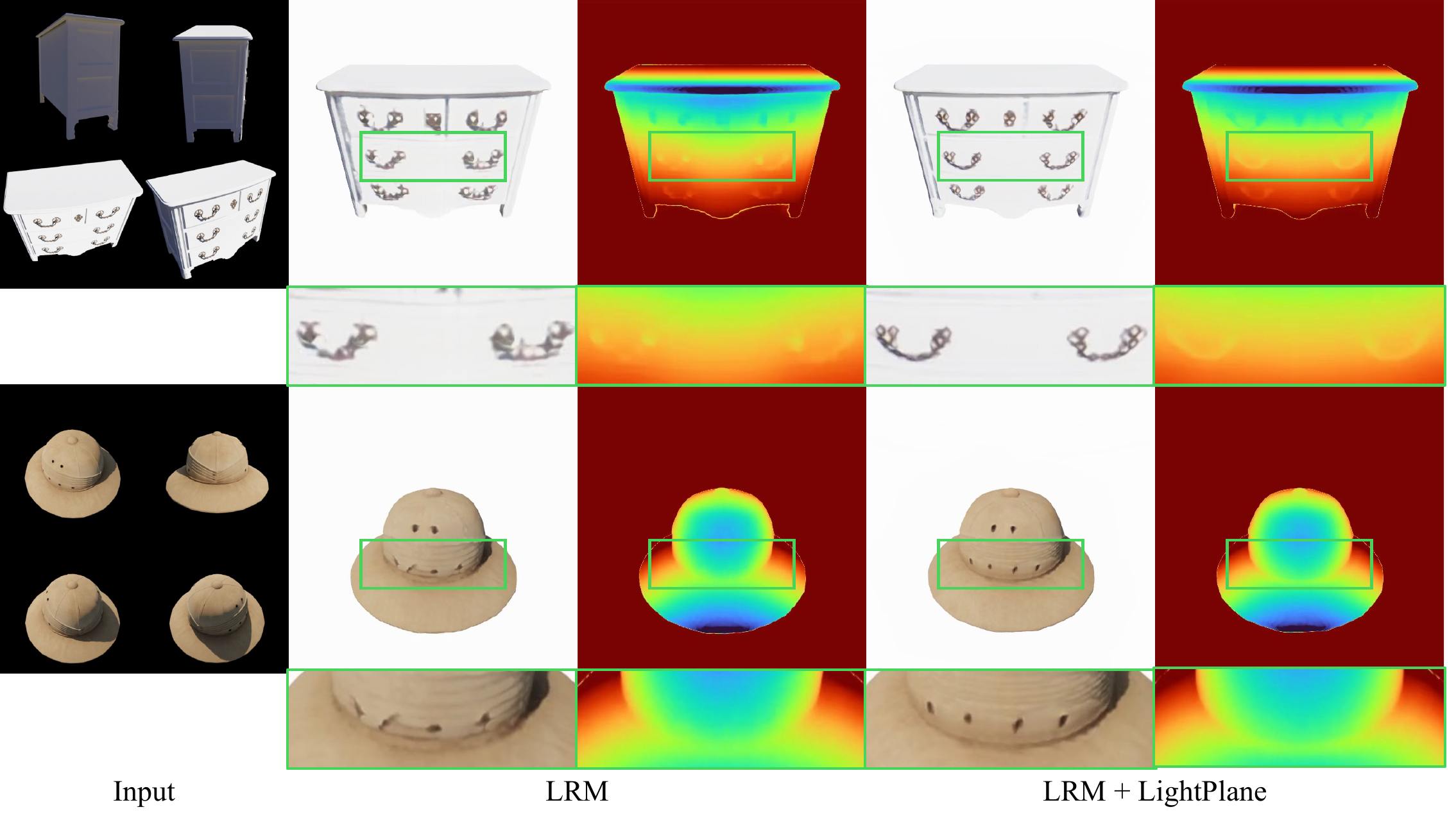}
    \caption{
    \textbf{Reconstruction comparison between LRM and LRM + \method}.
    }
    \label{fig:supp_lrm_com}
\end{figure}

%% file: figures/fig_sup_LRM.tex
\begin{figure*}[t]
\centering
\newcommand{\figw}{0.195\textwidth}
\renewcommand{\arraystretch}{0}
\setlength{\tabcolsep}{0pt}
\begin{tabular}{ccccc}
\includegraphics[width=\figw]{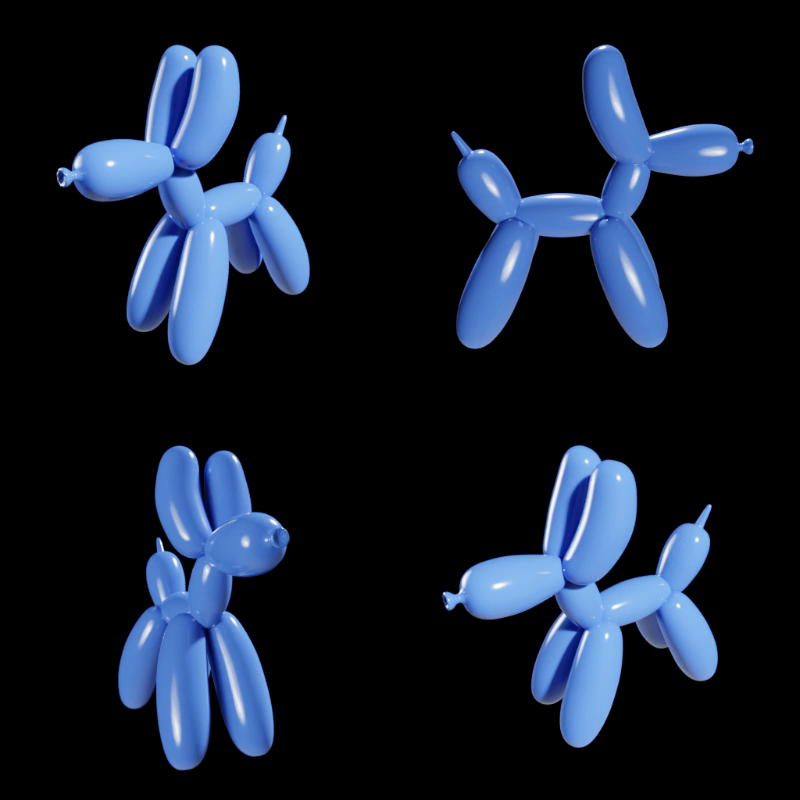}&
\includegraphics[width=\figw]{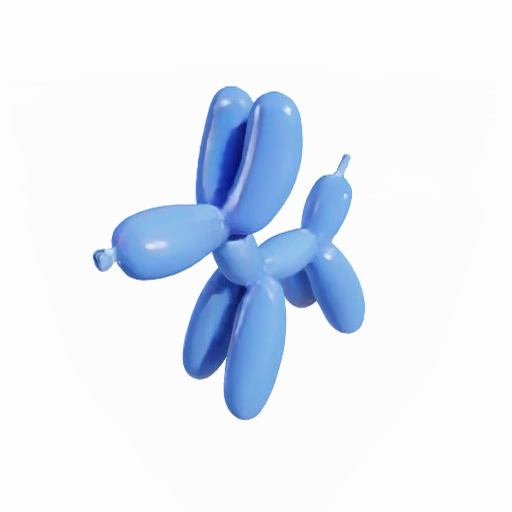}&
\includegraphics[width=\figw]{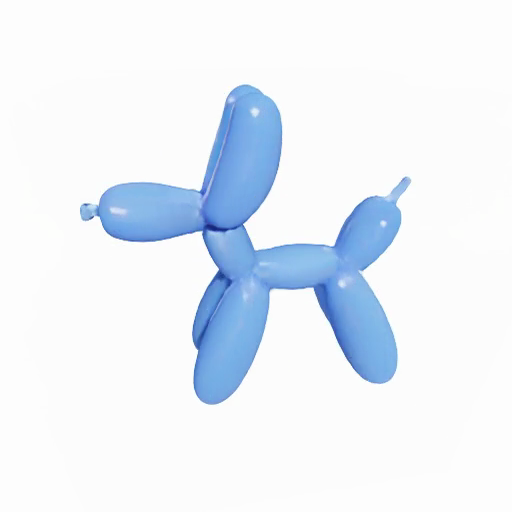}&
\includegraphics[width=\figw]{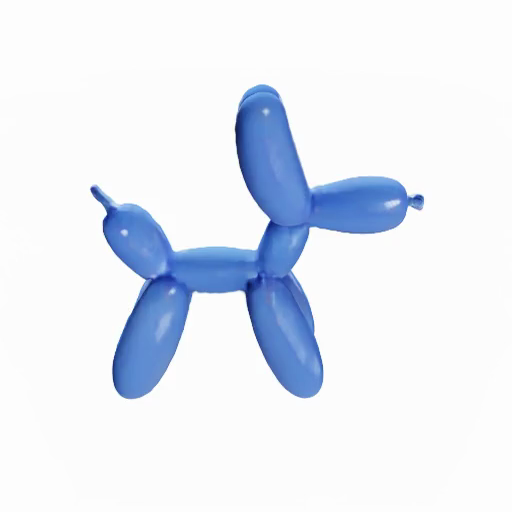}&
\includegraphics[width=\figw]{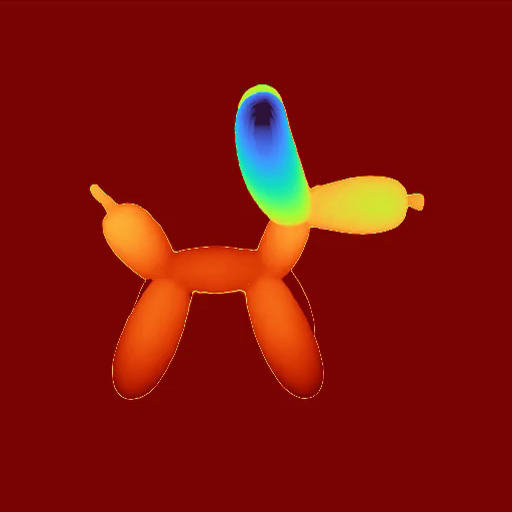}\\
\includegraphics[width=\figw]{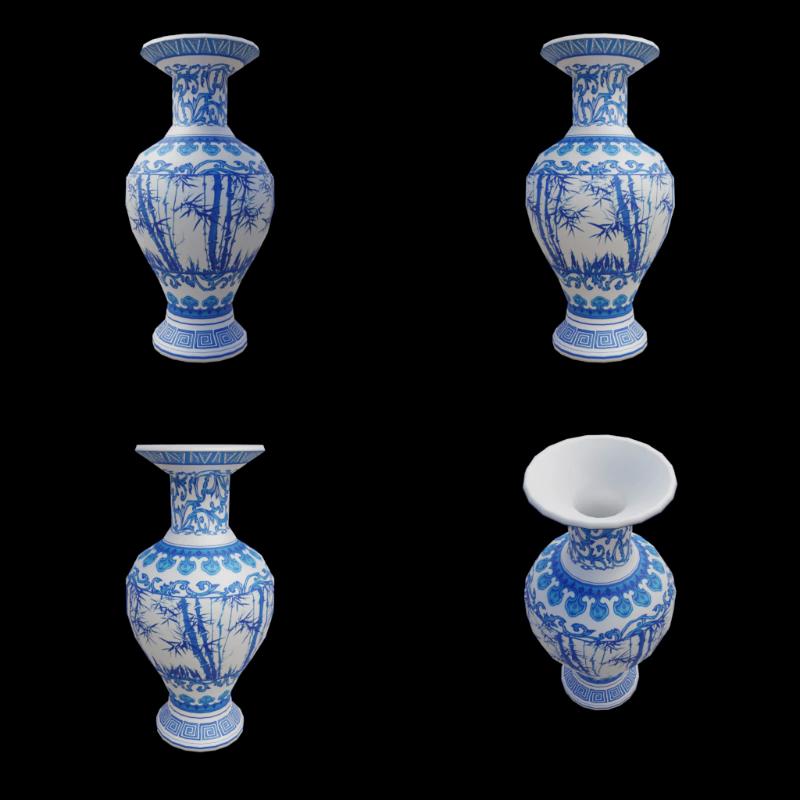}&
\includegraphics[width=\figw]{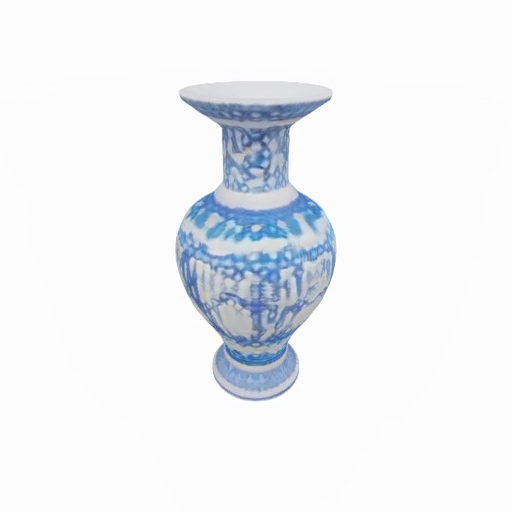}&
\includegraphics[width=\figw]{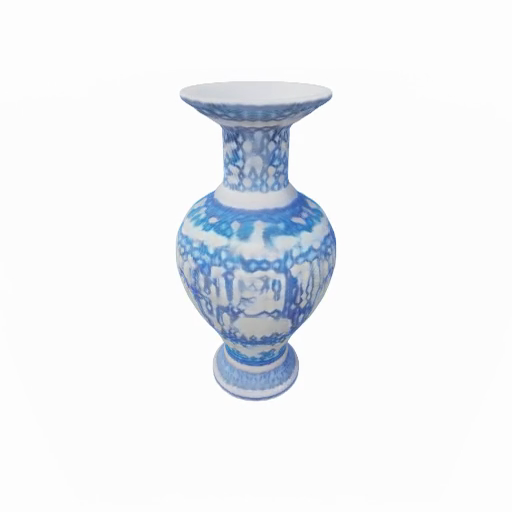}&
\includegraphics[width=\figw]{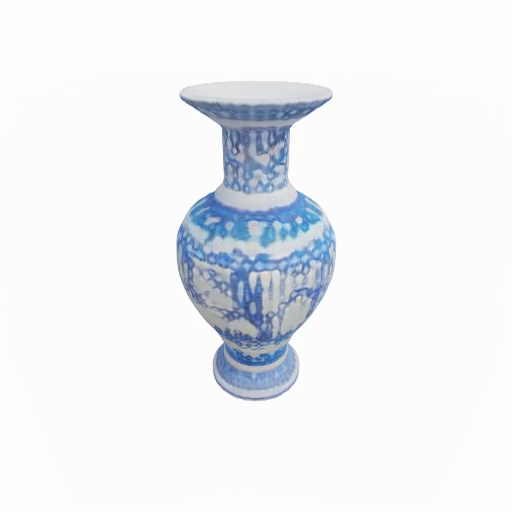}&
\includegraphics[width=\figw]{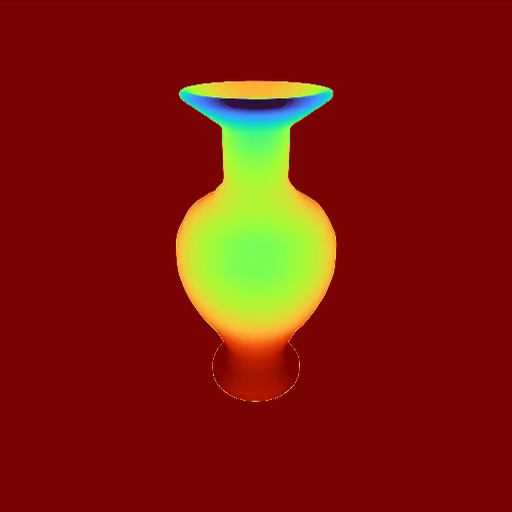}\\
\includegraphics[width=\figw]{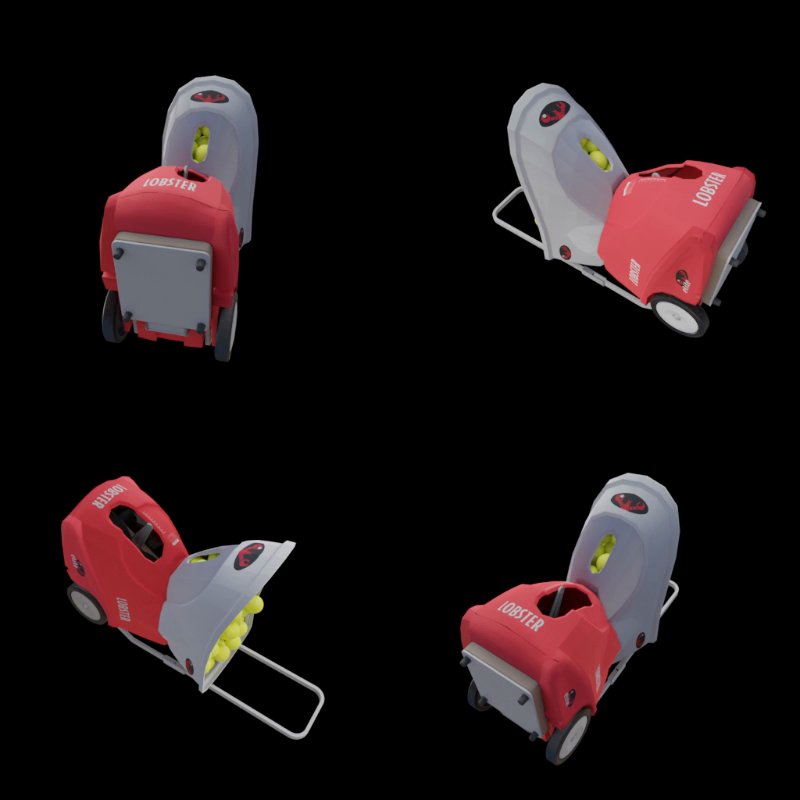}&
\includegraphics[width=\figw]{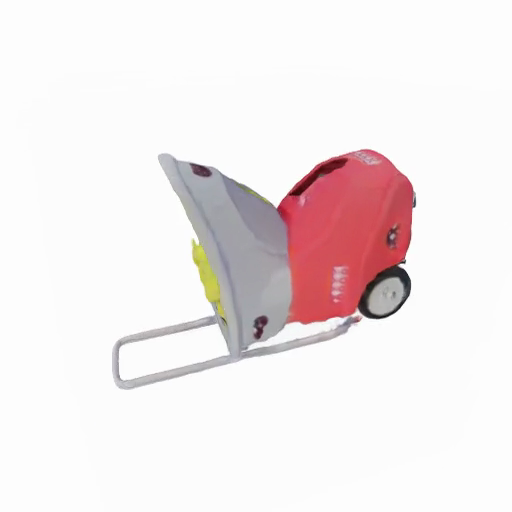}&
\includegraphics[width=\figw]{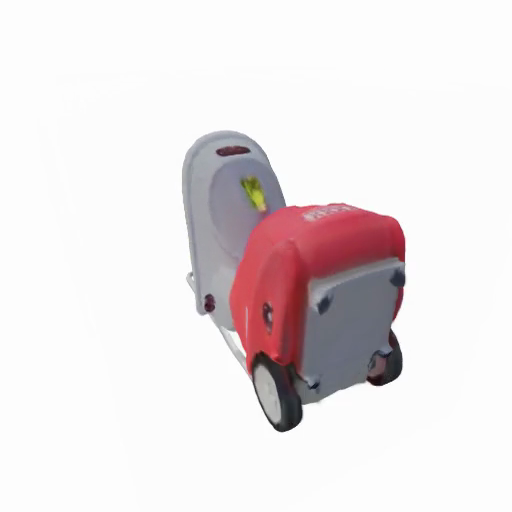}&
\includegraphics[width=\figw]{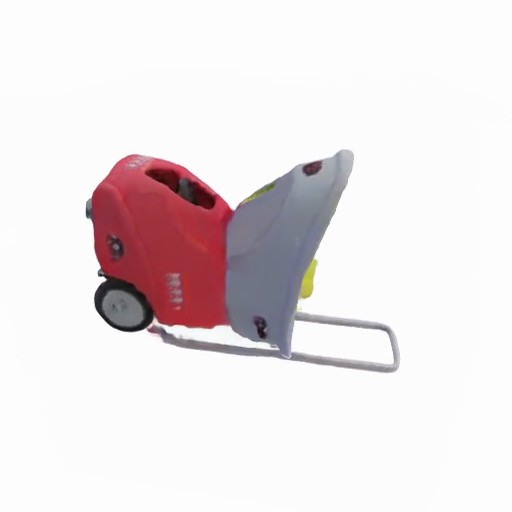}&
\includegraphics[width=\figw]{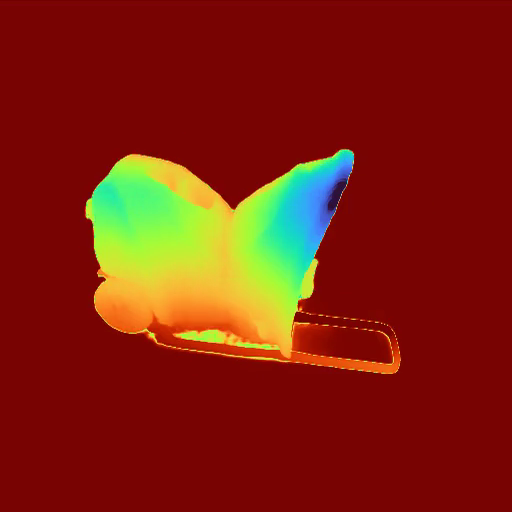}\\
\includegraphics[width=\figw]{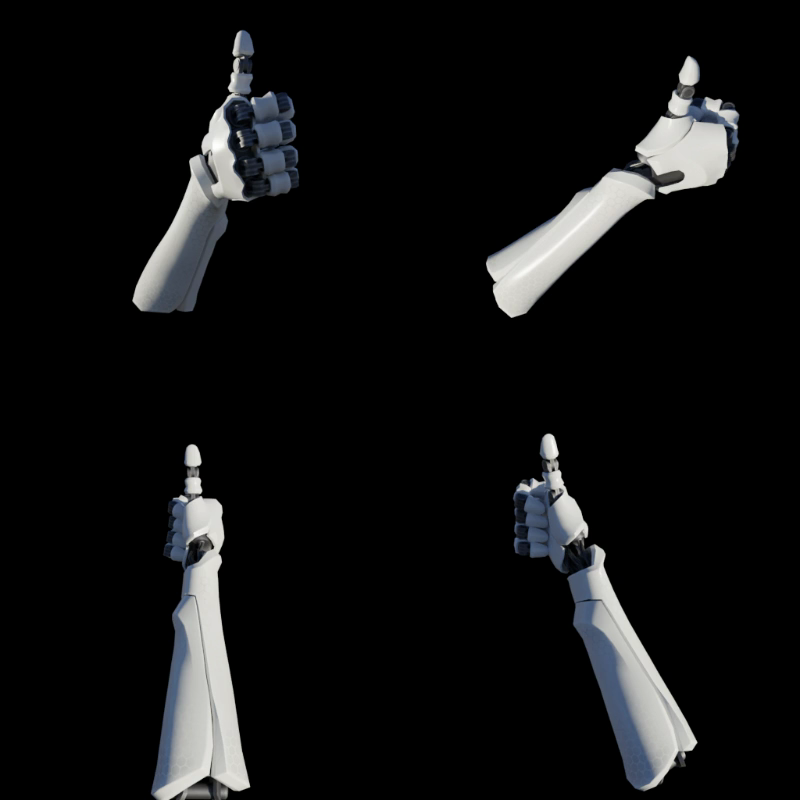}&
\includegraphics[width=\figw]{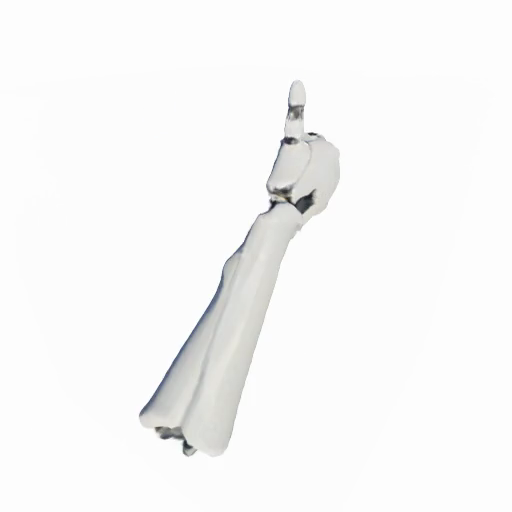}&
\includegraphics[width=\figw]{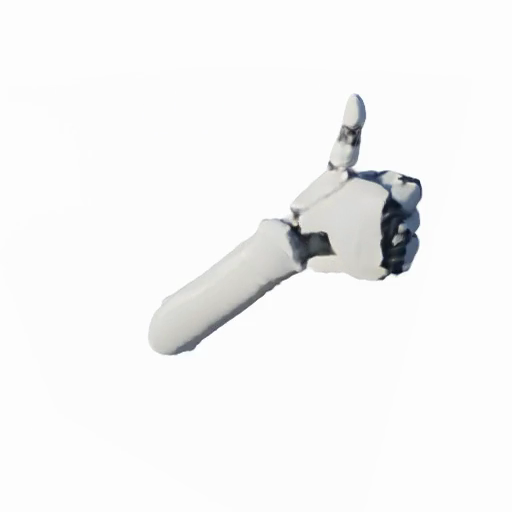}&
\includegraphics[width=\figw]{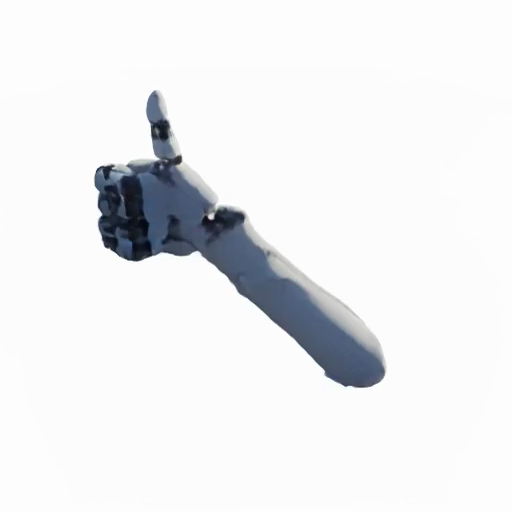}&
\includegraphics[width=\figw]{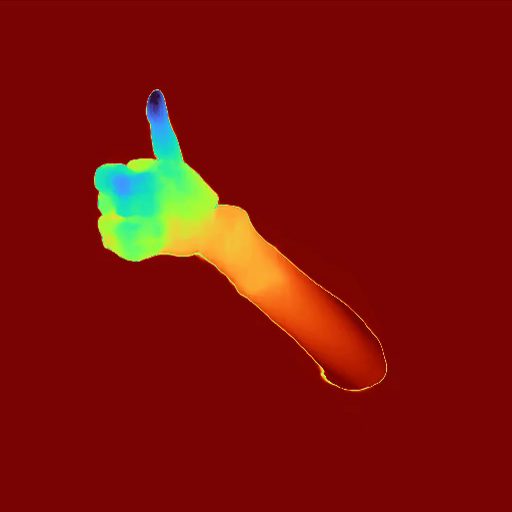}\\
\includegraphics[width=\figw]{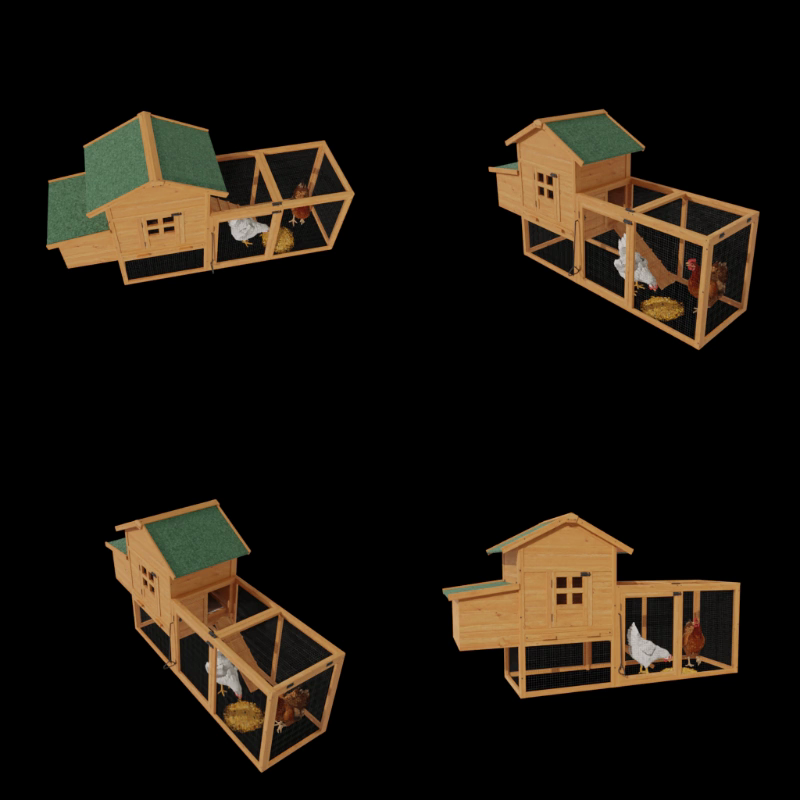}&
\includegraphics[width=\figw]{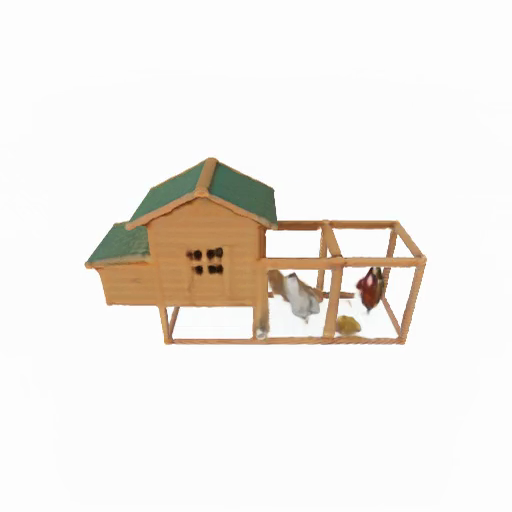}&
\includegraphics[width=\figw]{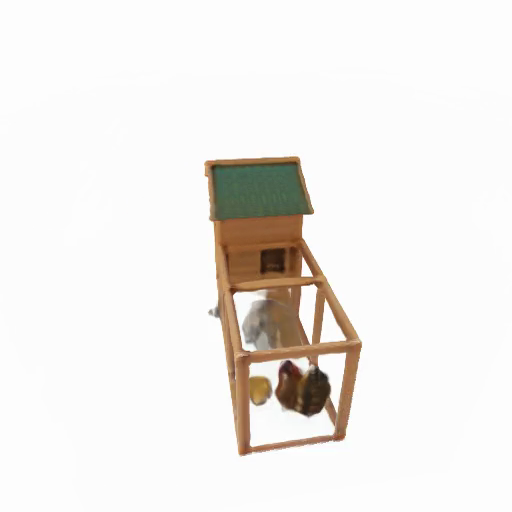}&
\includegraphics[width=\figw]{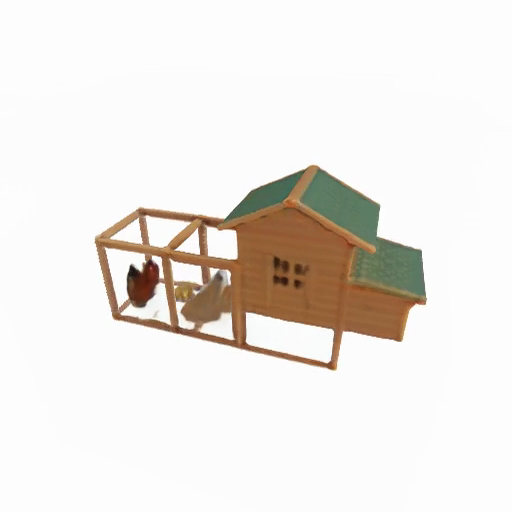}&
\includegraphics[width=\figw]{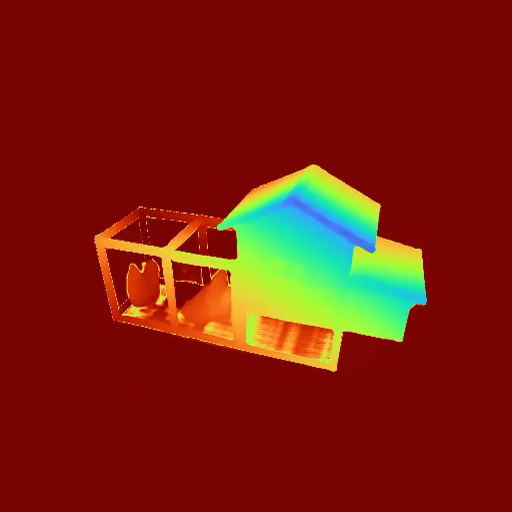}\\
\includegraphics[width=\figw]{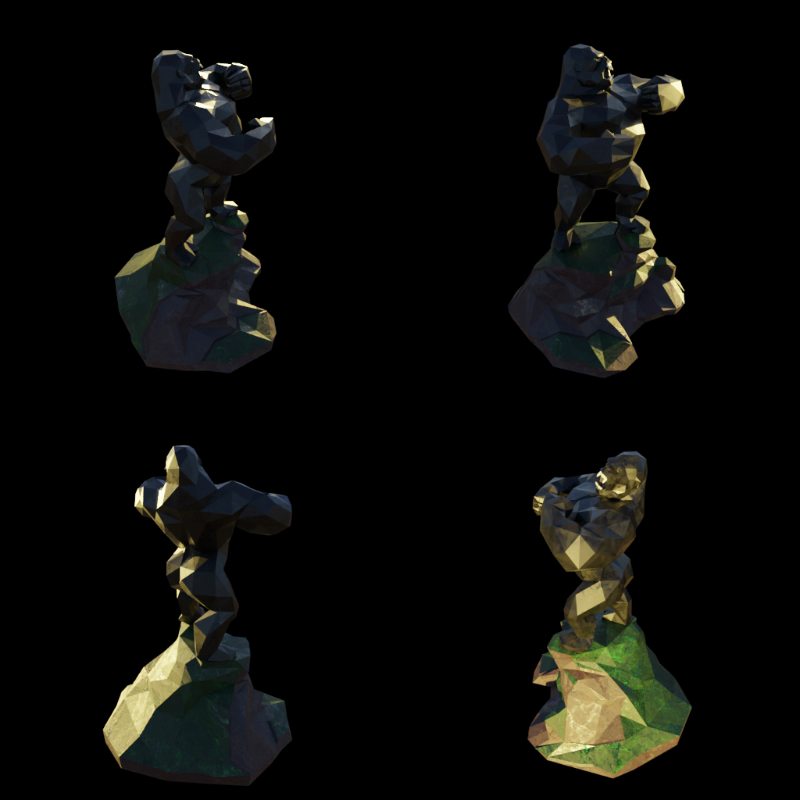}&
\includegraphics[width=\figw]{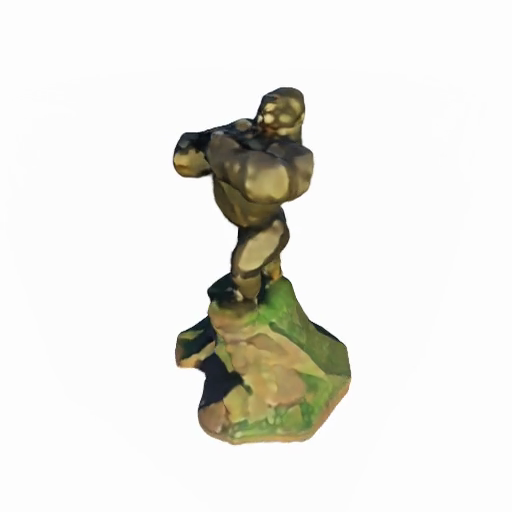}&
\includegraphics[width=\figw]{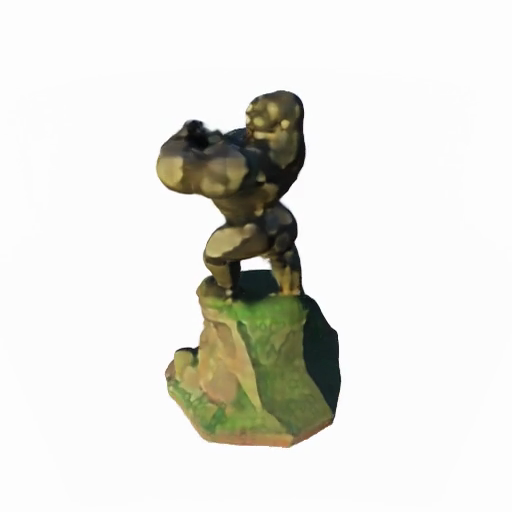}&
\includegraphics[width=\figw]{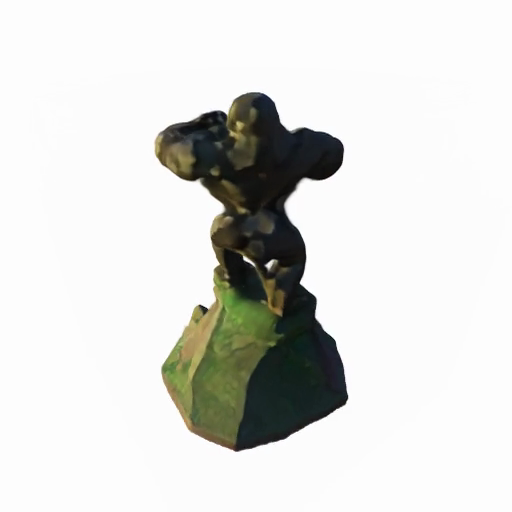}&
\includegraphics[width=\figw]{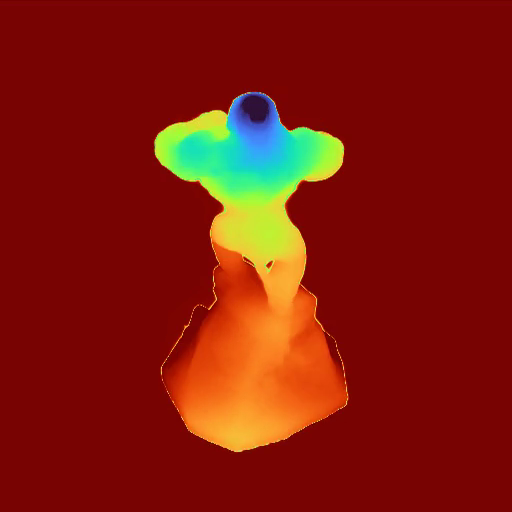}\\
\includegraphics[width=\figw]{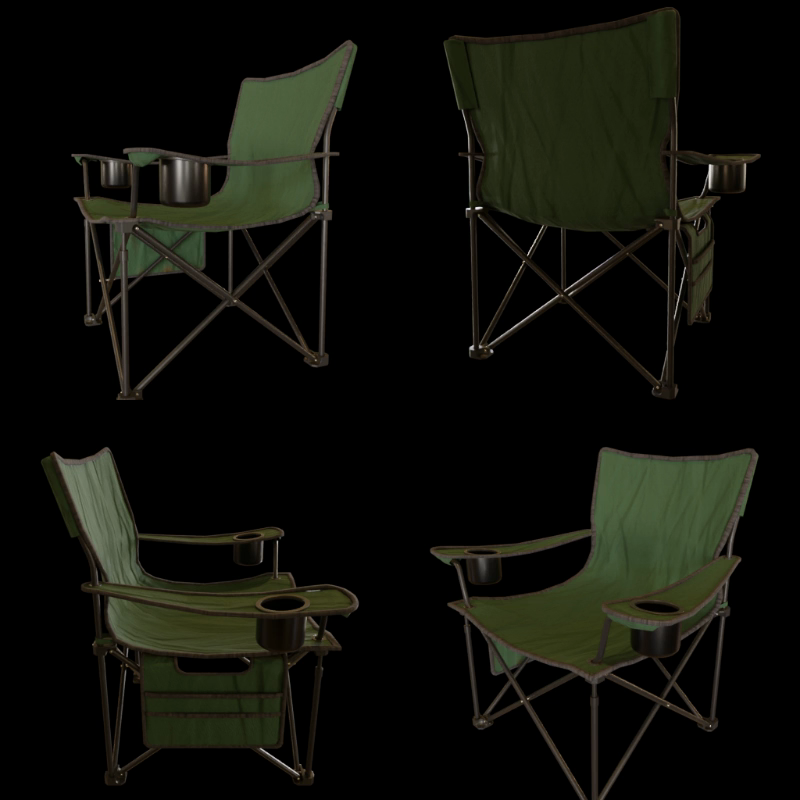}&
\includegraphics[width=\figw]{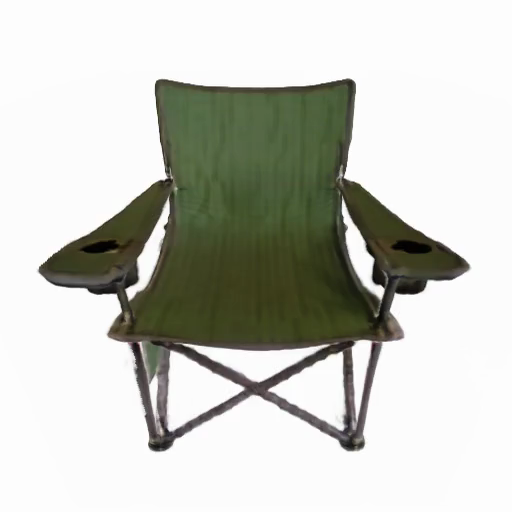}&
\includegraphics[width=\figw]{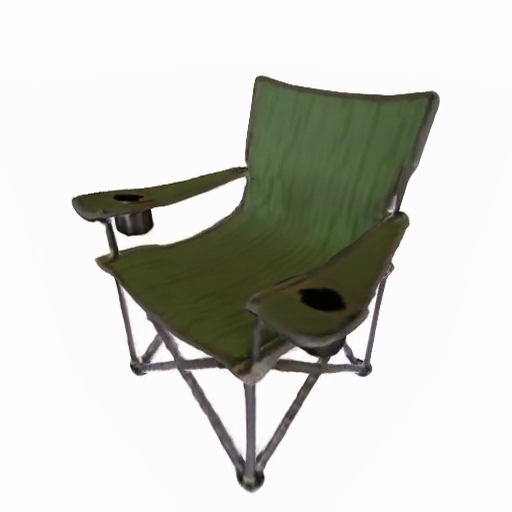}&
\includegraphics[width=\figw]{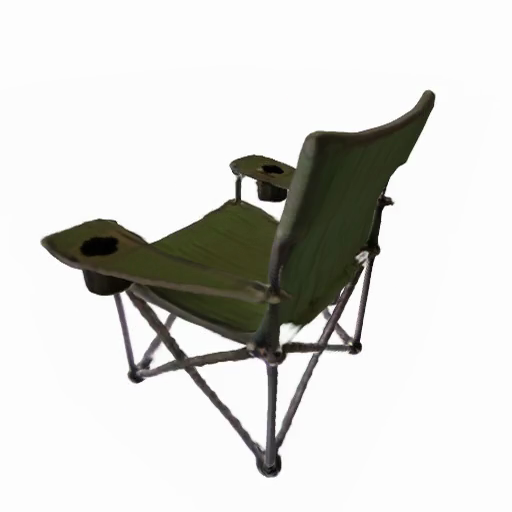}&
\includegraphics[width=\figw]{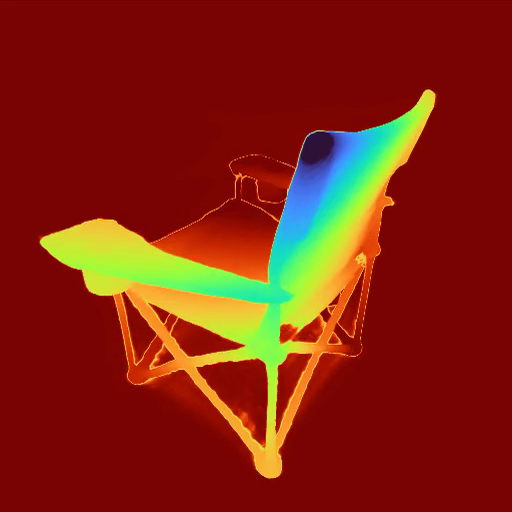}\\
\end{tabular}
\caption{%
\textbf{Multi-view LRM with \method}.
}
\label{fig:sup_lrm}
\end{figure*}

%% file: figures/fig_recon_depth.tex
\begin{figure}
\centering
\resizebox{\columnwidth}{!}{%
\footnotesize
\setlength\tabcolsep{0pt}
\begin{tabular}{ccccc}
    \text{\footnotesize Voxel}
    &\text{\footnotesize \method-\texttt{Vox}}
    &\text{\footnotesize TriPlane}
    &\text{\footnotesize \method-\texttt{Tri}}
    &\text{\footnotesize NeRF~\cite{mildenhall2020nerf}}
    \\
    \includegraphics[width=0.19\textwidth]{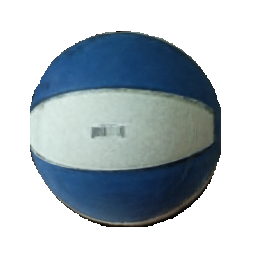}
    &\includegraphics[width=0.19\textwidth]{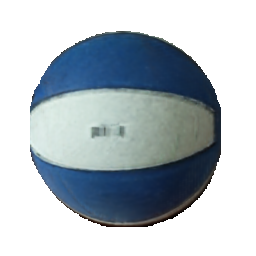}
    &\includegraphics[width=0.19\textwidth]{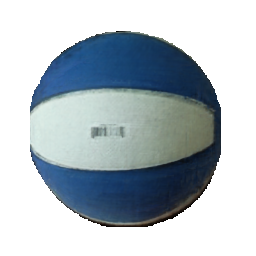}
    &\includegraphics[width=0.19\textwidth]{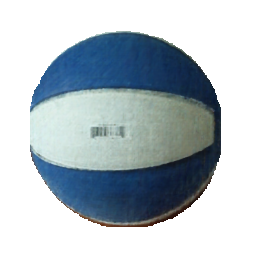}
    &\includegraphics[width=0.19\textwidth]{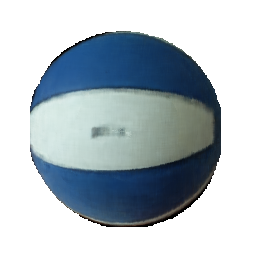}
    \\
    \includegraphics[width=0.19\textwidth]{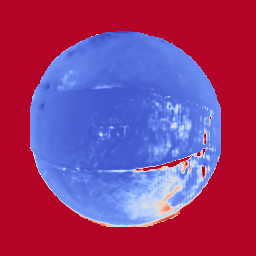}
    &\includegraphics[width=0.19\textwidth]{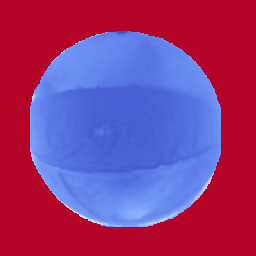}
    &\includegraphics[width=0.19\textwidth]{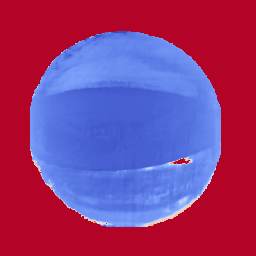}
    &\includegraphics[width=0.19\textwidth]{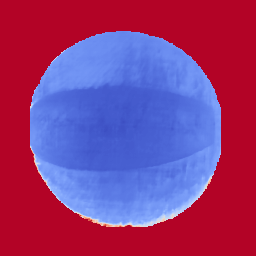}
    &\includegraphics[width=0.19\textwidth]{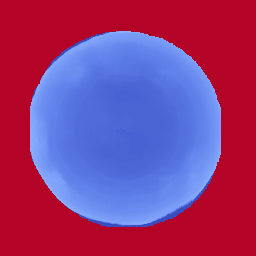}\\
    \text{0.373 / 35min}
    & \text{0.449 / 5min}
    & \text{0.492 / 27min}
    & \text{0.679 / 7min}
    & \text{ 0.658 / 1day}
\end{tabular}
}
\vspace{-3mm}
\caption{
\textbf{3D Reconstruction with Learned Initialization.} 
We show rendered images (top row) and depths.
Optimizing hashed representations (Voxel, Triplane) on real scenes leads to geometric defects.
Using our models (\method-\texttt{Vox}, \method-\texttt{Tri}), we first learn a reconstruction prior on CO3Dv2.
We then initialize reconstruction with a feed-forward pass accepting up to 100 source views of a single-scene.
After fine-tuning, we observe improved quality of the reconstructed geometry (columns 3 and 4).
We show \emph{Depth Corr.} ($\uparrow$) and \emph{Overfitting Time} ($\downarrow$) below images.
}%
\label{fig:3D_recon}
\end{figure}

%% file: figures/fig_sup_uncond.tex
\begin{figure*}[t]
\centering
\newcommand{\figw}{0.165\textwidth}
\renewcommand{\arraystretch}{0}
\setlength{\tabcolsep}{0pt}
\begin{tabular}{ccccccc}
\includegraphics[width=\figw]{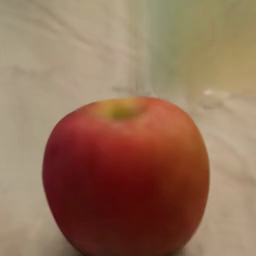}&
\includegraphics[width=\figw]{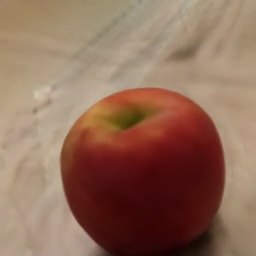}&
\includegraphics[width=\figw]{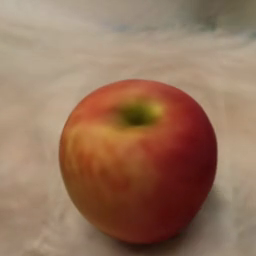}&
\includegraphics[width=\figw]{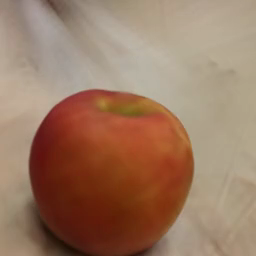}&
\includegraphics[width=\figw]{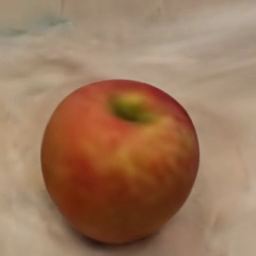}&
\includegraphics[width=\figw]{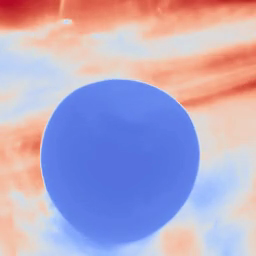}\\
\includegraphics[width=\figw]{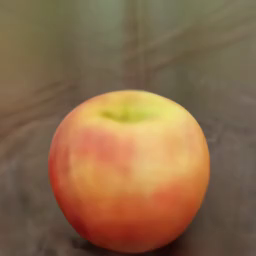}&
\includegraphics[width=\figw]{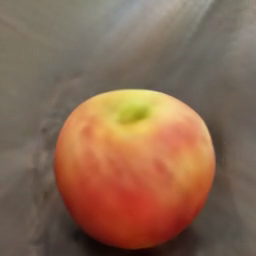}&
\includegraphics[width=\figw]{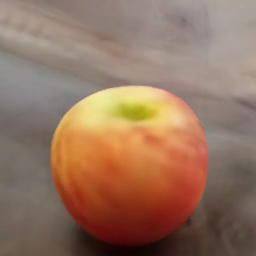}&
\includegraphics[width=\figw]{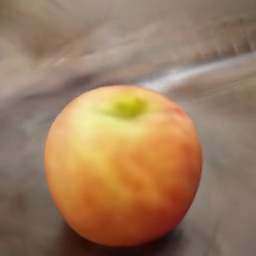}&
\includegraphics[width=\figw]{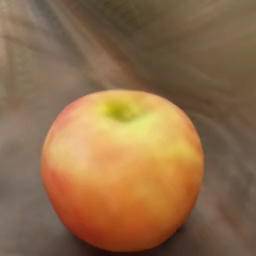}&
\includegraphics[width=\figw]{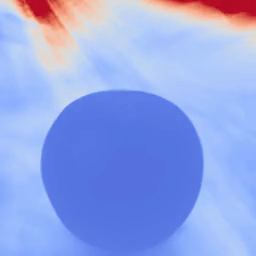}\\
\includegraphics[width=\figw]{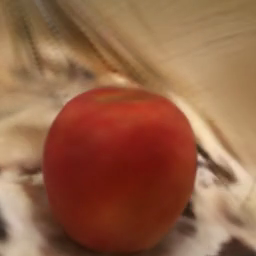}&
\includegraphics[width=\figw]{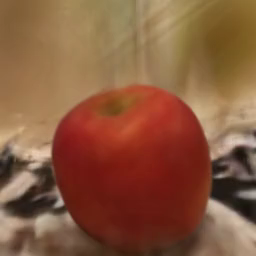}&
\includegraphics[width=\figw]{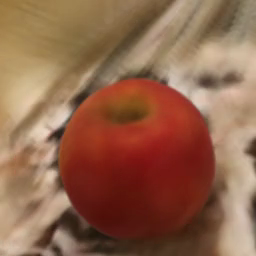}&
\includegraphics[width=\figw]{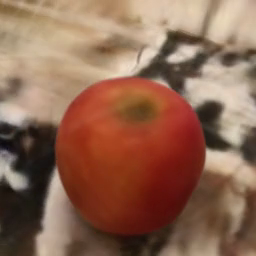}&
\includegraphics[width=\figw]{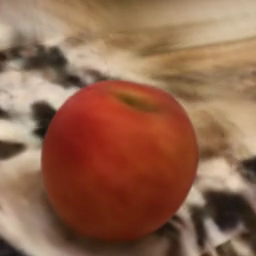}&
\includegraphics[width=\figw]{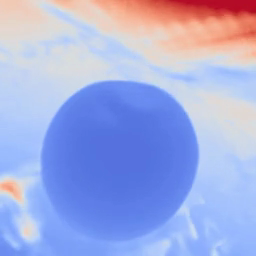}\\
\includegraphics[width=\figw]{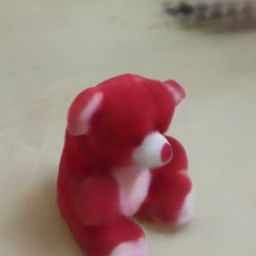}&
\includegraphics[width=\figw]{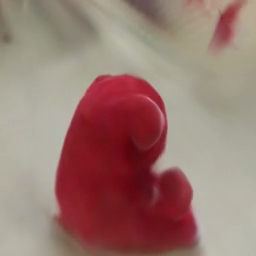}&
\includegraphics[width=\figw]{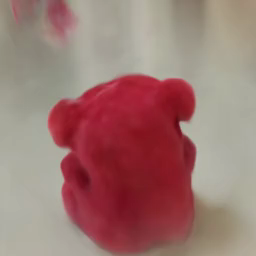}&
\includegraphics[width=\figw]{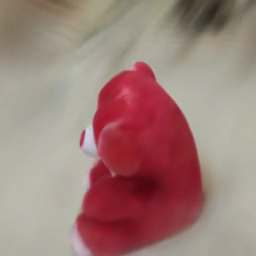}&
\includegraphics[width=\figw]{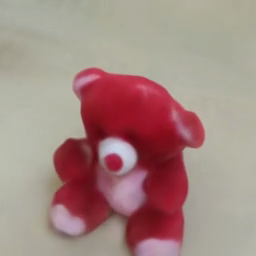}&
\includegraphics[width=\figw]{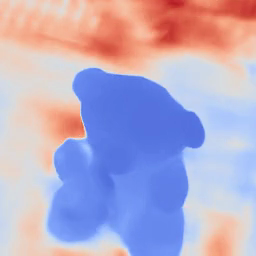}\\
\includegraphics[width=\figw]{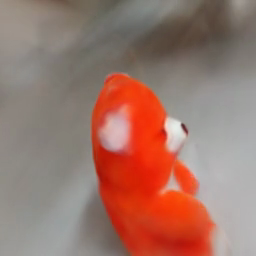}&
\includegraphics[width=\figw]{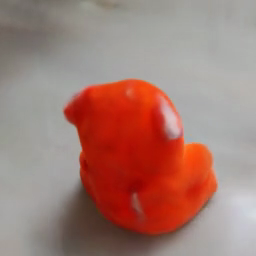}&
\includegraphics[width=\figw]{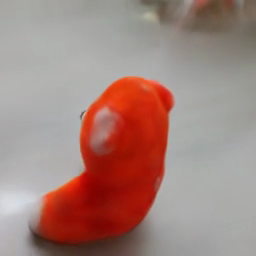}&
\includegraphics[width=\figw]{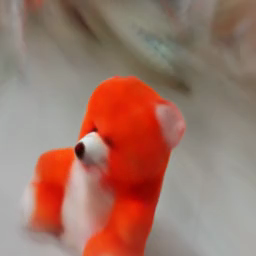}&
\includegraphics[width=\figw]{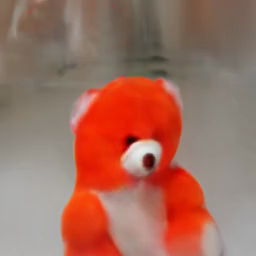}&
\includegraphics[width=\figw]{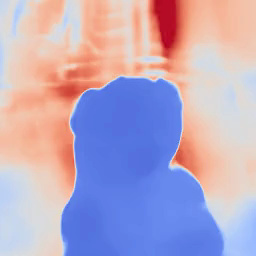}\\
\includegraphics[width=\figw]{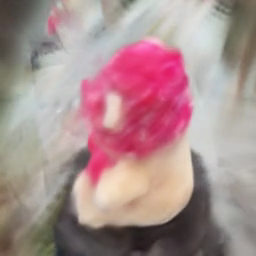}&
\includegraphics[width=\figw]{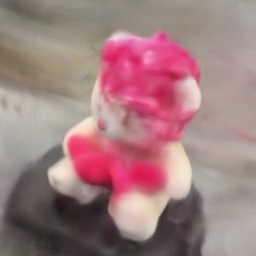}&
\includegraphics[width=\figw]{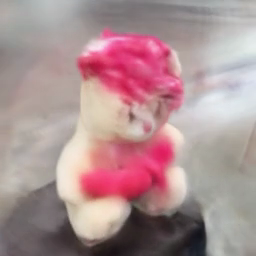}&
\includegraphics[width=\figw]{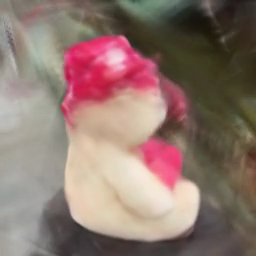}&
\includegraphics[width=\figw]{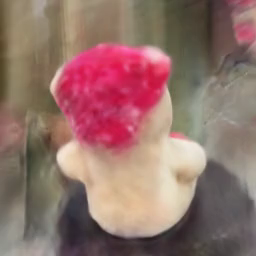}&
\includegraphics[width=\figw]{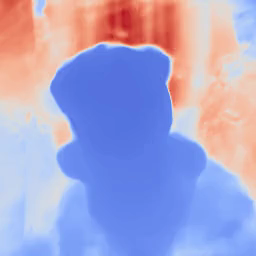}\\
\end{tabular}
\caption{%
\textbf{Unconditional 3D Generation}
displaying samples from our \method-augmented Viewset Diffusion trained on CO3Dv2~\cite{reizenstein2021common}.%
\label{fig:sup_uncond_gen_1}%
}
\end{figure*}

%% file: figures/fig_sup_uncond_2.tex
\begin{figure*}[t]
\centering
\newcommand{\figw}{0.165\textwidth}
\renewcommand{\arraystretch}{0}
\setlength{\tabcolsep}{0pt}
\begin{tabular}{ccccccc}
\includegraphics[width=\figw]{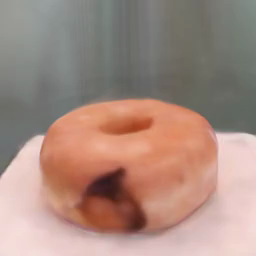}&
\includegraphics[width=\figw]{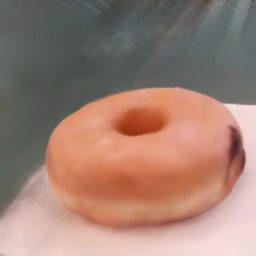}&
\includegraphics[width=\figw]{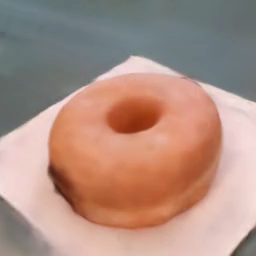}&
\includegraphics[width=\figw]{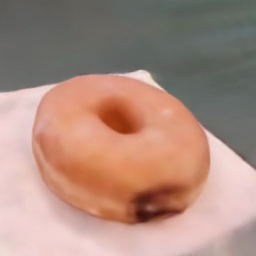}&
\includegraphics[width=\figw]{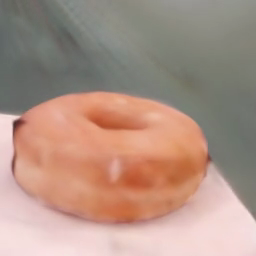}&
\includegraphics[width=\figw]{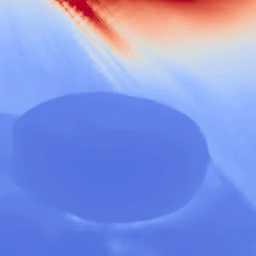}\\
\includegraphics[width=\figw]{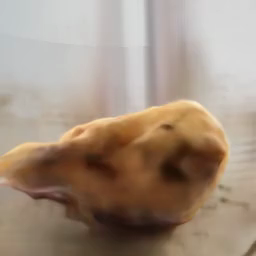}&
\includegraphics[width=\figw]{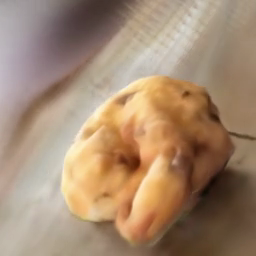}&
\includegraphics[width=\figw]{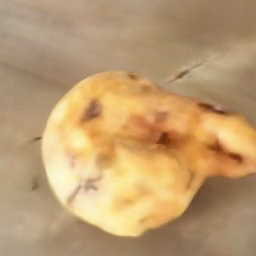}&
\includegraphics[width=\figw]{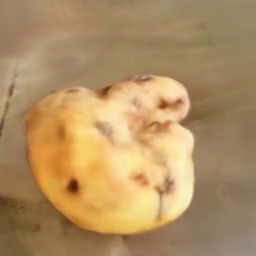}&
\includegraphics[width=\figw]{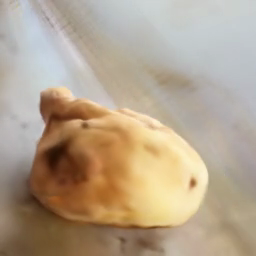}&
\includegraphics[width=\figw]{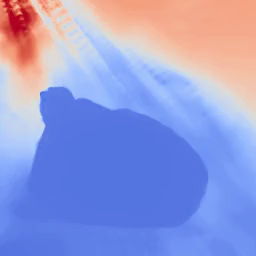}\\
\includegraphics[width=\figw]{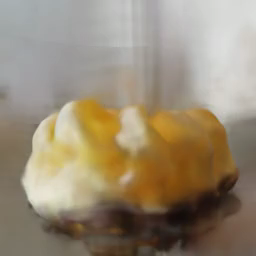}&
\includegraphics[width=\figw]{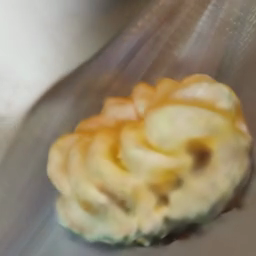}&
\includegraphics[width=\figw]{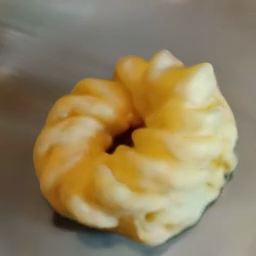}&
\includegraphics[width=\figw]{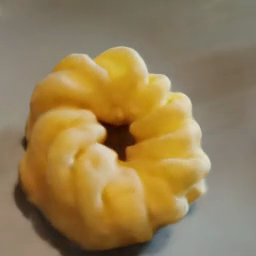}&
\includegraphics[width=\figw]{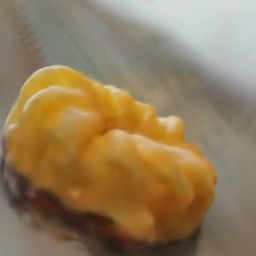}&
\includegraphics[width=\figw]{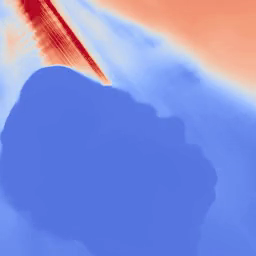}\\
\includegraphics[width=\figw]{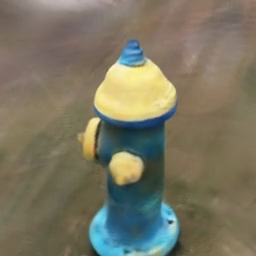}&
\includegraphics[width=\figw]{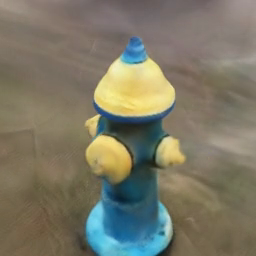}&
\includegraphics[width=\figw]{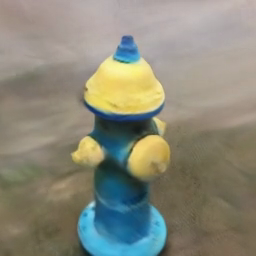}&
\includegraphics[width=\figw]{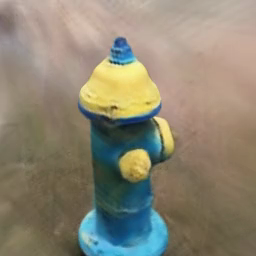}&
\includegraphics[width=\figw]{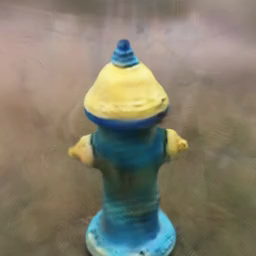}&
\includegraphics[width=\figw]{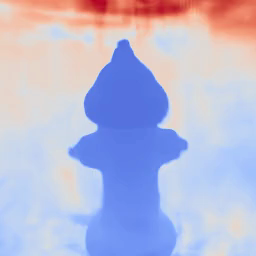}\\
\includegraphics[width=\figw]{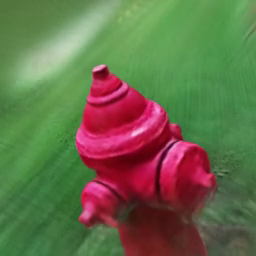}&
\includegraphics[width=\figw]{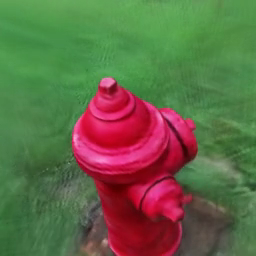}&
\includegraphics[width=\figw]{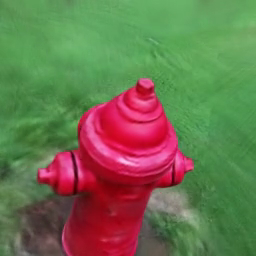}&
\includegraphics[width=\figw]{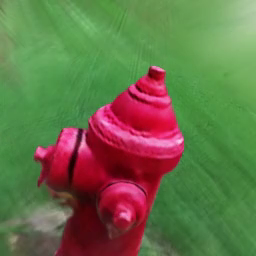}&
\includegraphics[width=\figw]{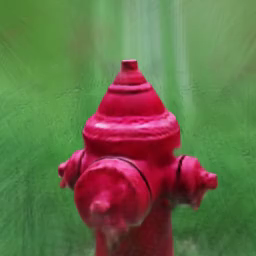}&
\includegraphics[width=\figw]{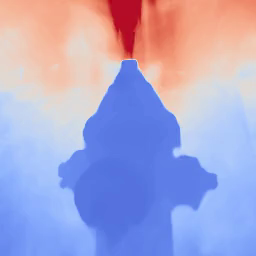}\\
\includegraphics[width=\figw]{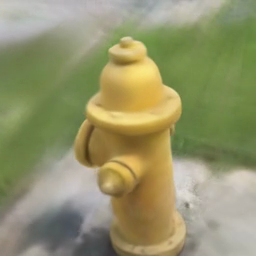}&
\includegraphics[width=\figw]{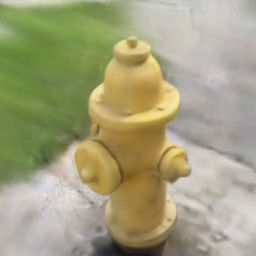}&
\includegraphics[width=\figw]{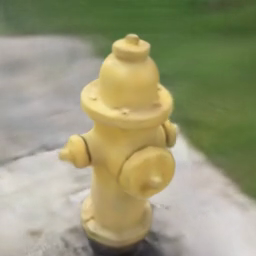}&
\includegraphics[width=\figw]{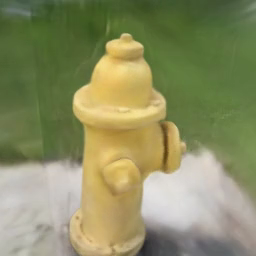}&
\includegraphics[width=\figw]{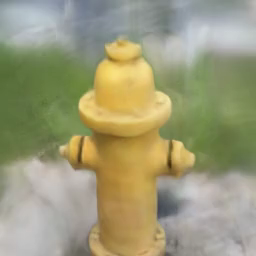}&
\includegraphics[width=\figw]{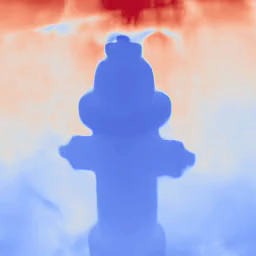}\\
\end{tabular}
\caption{%
\textbf{Unconditional 3D Generation}
displaying samples from our \method-augmented Viewset Diffusion trained on CO3Dv2~\cite{reizenstein2021common}.%
\label{fig:sup_uncond_gen_2}%
}
\end{figure*}

%% file: figures/fig_cond_gen.tex
\begin{figure}
\centering
\setlength\tabcolsep{1pt}
    \begin{tabular}{c|ccc}
    \multicolumn{1}{c}{Input View} 
    &\multicolumn{3}{c}{Novel Views} \\
    \includegraphics[height=0.23\textwidth]{figures/materials/cond_gen/cond_1_crop.jpg}  &\includegraphics[width=0.23\textwidth]{figures/materials/cond_gen/1_1.png}
    &\includegraphics[width=0.23\textwidth]{figures/materials/cond_gen/1_2.png}  &\includegraphics[width=0.23\textwidth]{figures/materials/cond_gen/1_3.png}\\ 
  \includegraphics[height=0.23\textwidth]{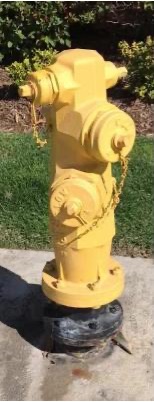}  &\includegraphics[width=0.23\textwidth]{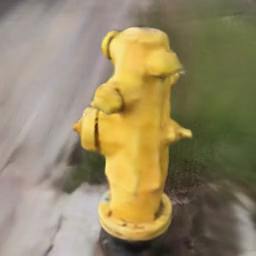}
    &\includegraphics[width=0.23\textwidth]{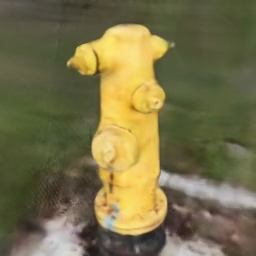}  &\includegraphics[width=0.23\textwidth]{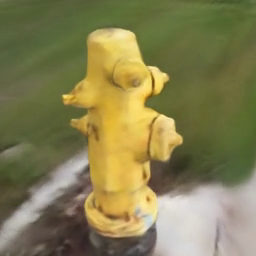}\\   
    
    \end{tabular}
\caption{\textbf{Monocular 3D Reconstruction on CO3Dv2.} With a single clean image as input, our model could generate realistic 3D structures matching the input views. }
    \label{fig:single_view_recon}
\end{figure}

%% file: figures/fig_text_3D_co3d.tex
\begin{figure*}
\centering
\newcommand{\textthreedfigwidth}{0.23\textwidth}
\setlength\tabcolsep{1pt}
\begin{tabular}{cccc}%
\includegraphics[width=\textthreedfigwidth]{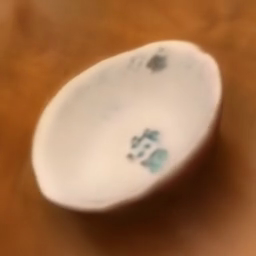}%
&\includegraphics[width=\textthreedfigwidth]{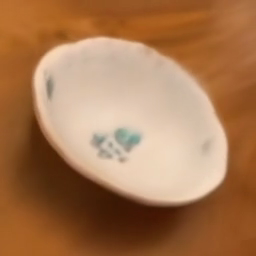}%
&\includegraphics[width=\textthreedfigwidth]{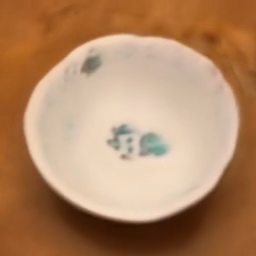}%
&\includegraphics[width=\textthreedfigwidth]{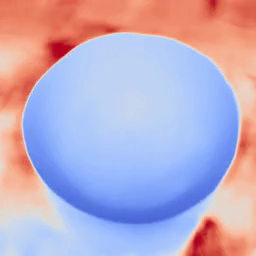}\\
\multicolumn{4}{c}{\textit{A white bowl with a blue and white fish in the center}} \\
\includegraphics[width=\textthreedfigwidth]{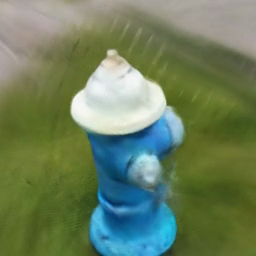}%
&\includegraphics[width=\textthreedfigwidth]{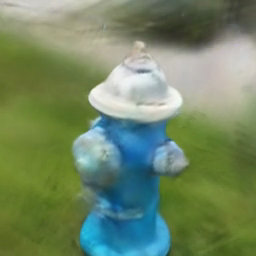}%
&\includegraphics[width=\textthreedfigwidth]{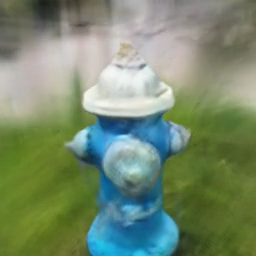}%
&\includegraphics[width=\textthreedfigwidth]{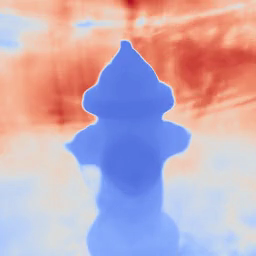}\\
\multicolumn{4}{c}{\textit{A blue and white fire hydrant fire hydrant in a grassy area}} \\
\includegraphics[width=\textthreedfigwidth]{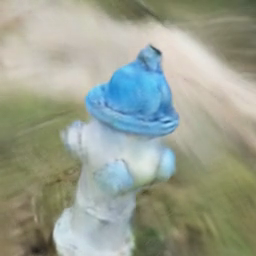}%
&\includegraphics[width=\textthreedfigwidth]{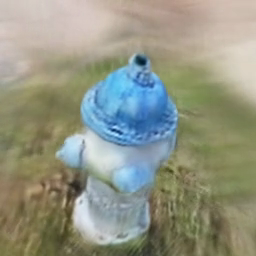}%
&\includegraphics[width=\textthreedfigwidth]{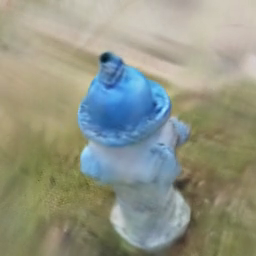}%
&\includegraphics[width=\textthreedfigwidth]{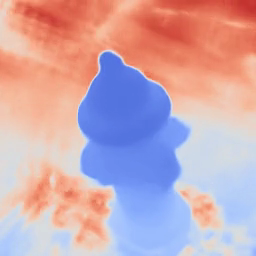}\\
\multicolumn{4}{c}{\textit{A blue and white fire hydrant with a blue cap on the top}}
\end{tabular}
    \caption{\textbf{Text-Conditioned Generation on CO3Dv2.}
    Our pipeline could generate 3D structures with text input as conditions. 
    }
    \label{fig:text_3D_co3d}
    \vspace{-4mm}
\end{figure*}